\documentclass{article}
\usepackage{iclr2027_conference,times}

\usepackage{amsmath,amsfonts,bm}

\def\eqref#1{equation~\ref{#1}}

\def\1{\bm{1}}

\DeclareMathAlphabet{\mathsfit}{\encodingdefault}{\sfdefault}{m}{sl}
\SetMathAlphabet{\mathsfit}{bold}{\encodingdefault}{\sfdefault}{bx}{n}

\DeclareMathOperator*{\argmin}{arg\,min}

\usepackage{float}
\usepackage{xcolor}
\definecolor{darkblue}{RGB}{0,0,139}
\definecolor{linkgreen}{RGB}{44,160, 44}
\usepackage{hyperref}
\hypersetup{
  colorlinks=true,
  linkcolor=darkblue,
  citecolor=linkgreen!70!black,
  urlcolor=linkgreen!70!black
}
\usepackage{url}

\usepackage{amsmath, amssymb, amsfonts}
\usepackage[nameinlink,noabbrev,capitalize]{cleveref}

\usepackage{booktabs}
\usepackage{multirow}

\usepackage{pgfplots}
\pgfplotsset{compat=1.17}

\usepackage{algorithm}
\usepackage{algpseudocode}
\newcommand{\algrule}[1][.4pt]{\par\vskip.3\baselineskip\hrule height #1\par\vskip.3\baselineskip}

\usepackage{glossaries}
\glsdisablehyper
\newacronym{llm}{LLM}{Large Language Model}
\newacronym{gpt}{GPT}{Generative Pretrained Transformer}
\newacronym{nlp}{NLP}{Natural Language Processing}
\newacronym{bn}{BN}{Batch-Normalization}
\newacronym{ln}{LN}{Layer-Normalization}
\newacronym{obs}{OBS}{Optimal Brain Surgeon}
\newacronym{obd}{OBD}{Optimal Brain Damage}
\newacronym{kv}{KV}{Key-Value}

\usepackage{pifont}
\newcommand{\cmark}{\ding{51}}
\newcommand{\xmark}{\ding{55}}

\newcommand{\diag}{\operatorname{diag}}
\newcommand{\tr}{\operatorname{tr}}
\newcommand{\one}{\mathbf{1}}

\newtheorem{lemma}{Lemma}

\newcommand{\se}[1]{\,{\scriptsize\textcolor{black!55}{$\pm$#1}}}

\definecolor{cobs}{RGB}{ 44,160, 44}
\definecolor{cam}{RGB}{255,127, 14}
\definecolor{comp}{RGB}{227,119,194}
\definecolor{cloo}{RGB}{ 63, 81,181}
\definecolor{cobc}{RGB}{ 23,190,207}
\definecolor{ch2o}{RGB}{140, 86, 75}
\definecolor{ckvm}{RGB}{178, 34, 34}
\definecolor{ckvz}{RGB}{128,128,  0}
\definecolor{csum}{RGB}{ 40, 40, 40}
\definecolor{cext}{RGB}{127,127,127}
\definecolor{cmag}{RGB}{ 44,160, 44}

\pgfplotsset{
  ratio panel/.style={
    xlabel={Compaction ratio},
    xtick={0.01,0.02,0.05,0.1,0.2},
    xticklabels={$100\times$,$50\times$,$20\times$,$10\times$,$5\times$},
    xmin=0.008, xmax=0.26,
    grid=both,
    grid style={gray!20},
    tick label style={font=\scriptsize},
    label style={font=\small},
    title style={font=\small},
    legend cell align=left,
    legend style={font=\scriptsize, fill=white, fill opacity=0.9,
                  text opacity=1, draw=gray!40},
    every mark/.append style={solid},
    mark size=2pt,
    line width=0.9pt,
  },
  lobs/.style ={color=cobs!85!black, mark=triangle*, mark size=2.5pt},
  lobsf/.style={color=cobs!70, mark=triangle, mark size=2.5pt},
  lam/.style  ={color=cam,  mark=pentagon*},
  lamf/.style ={color=cam!65, mark=pentagon},
  lomp/.style ={color=comp!85!black, mark=diamond*, dashed, mark size=2.4pt},
  lompf/.style={color=comp!70, mark=diamond, dashed, mark size=2.4pt},
  lloo/.style ={color=cloo, mark=*},
  lloof/.style={color=cloo!65, mark=o},
  lobc/.style ={color=cobc, mark=square},
  lh2of/.style={color=ch2o!75, mark=o},
  lh2o/.style ={color=ch2o, mark=*},
  lkvmf/.style={color=ckvm!75, mark=Mercedes star},
  lkvm/.style ={color=ckvm, mark=Mercedes star},
  lkvzf/.style={color=ckvz!75, mark=oplus},
  lkvzs/.style={color=ckvz, mark=oplus*},
  lsum/.style ={color=csum, mark=star, only marks, mark size=3pt},
  ll2/.style  ={color=cext, mark=square},
  lqf/.style  ={color=cext!60!black, mark=x},
  lstr/.style ={color=cext!80!black, mark=+},
  lkv/.style  ={color=cmag, mark=square*},
}

\pgfplotsset{
  intro frontier panel/.style={
    xlabel={Pipeline time per context (seconds, log scale)},
    xmin=3, xmax=7000,
    ymin=24, ymax=86,
    grid=both,
    grid style={gray!20},
    tick label style={font=\small},
    label style={font=\small},
    title style={font=\small},
    every mark/.append style={solid},
  },
  fullctx/.style={black, dashed, thick, forget plot},
  pobs/.style ={only marks, mark=triangle*, mark size=3.8pt, color=cobs},
  pobs2/.style={only marks, mark=triangle,  mark size=3.8pt, color=cobs!70, line width=1.1pt},
  pam1/.style ={only marks, mark=pentagon*, mark size=3.5pt, color=cam},
  pam2/.style ={only marks, mark=pentagon,  mark size=3.5pt, color=cam!70, line width=1.1pt},
  pomps/.style={only marks, mark=diamond*,  mark size=3.8pt, color=comp!85!black},
  pomp1/.style={only marks, mark=diamond,   mark size=3.8pt, color=comp!70, line width=1.1pt},
  ploo1/.style={only marks, mark=*,         mark size=3pt, color=cloo},
  ploo2/.style={only marks, mark=o,         mark size=3pt, color=cloo!70, line width=1.1pt},
  pobc/.style ={only marks, mark=square,    mark size=2.9pt, color=cobc, line width=1.1pt},
  ph2o/.style ={only marks, mark=o,         mark size=3pt, color=ch2o, line width=1.1pt},
  pkvm/.style ={only marks, mark=Mercedes star, mark size=3.6pt, color=ckvm, line width=1.1pt},
  pkvz/.style ={only marks, mark=oplus,     mark size=3pt, color=ckvz, line width=1.1pt},
  pkvzs/.style={only marks, mark=oplus*,    mark size=3pt, color=ckvz, line width=1pt},
  psum/.style ={only marks, mark=star,      mark size=3.6pt, color=csum, line width=1pt},
  pmag/.style ={only marks, mark=square*,   mark size=3pt, color=cmag},
  pl2/.style  ={only marks, mark=square,    mark size=2.6pt, color=cext, line width=1pt},
  pqf/.style  ={only marks, mark=x,         mark size=3.2pt, color=cext!60!black, line width=1pt},
  pstr/.style ={only marks, mark=+,         mark size=3.2pt, color=cext!80!black, line width=1pt},
  fnew/.style ={mark=none, black, line width=1.6pt},
  fold/.style ={mark=none, gray!70!black, line width=1.1pt, dashdotted},
  ebar/.style={
    error bars/y dir=both, error bars/y explicit,
    error bars/error bar style={line width=0.9pt},
    error bars/error mark options={rotate=90, mark size=2.6pt, line width=0.9pt},
  },
}

\title{The \textsc{KV-Surgeon}: Compensation-Aware Offline KV-Cache Compaction}

\author{
  Moritz Wagner$^1$\thanks{Correspondence to: Moritz Wagner \ \texttt{wagner@zib.de}} \quad Christophe Roux$^{1\ 2}$ \quad Max Zimmer$^{1\ 2}$ \quad Sebastian Pokutta$^{1\ 2}$\\
  $^1$Department for AI in Society, Science, and Technology, Zuse Institute Berlin, Germany\\
  $^2$Institute of Mathematics, Technische Universität Berlin, Germany
}

\iclrfinalcopy

\begin{document}

\maketitle
\lhead{Preprint. Under review.}

\begin{abstract}
   Offline compaction of the key--value cache (KV-cache) reduces the cost of repeatedly querying a fixed context by constructing a smaller cache once and reusing it subsequently. The central challenge of compaction is deciding which cache entries to delete and which to retain. Existing methods score each candidate for deletion on its own, with the rest of the cache being held fixed. However, softmax renormalization of the attention operation couples these decisions in a nonlinear manner: removing one entry changes the attention weights of every remaining one. In this work, we introduce \textsc{KV-Surgeon}, which uses calibration data to take interactions into account in a tractable way. We derive a quadratic surrogate that admits closed-form expressions for jointly making the optimal greedy removal decision as well as compensation within the remaining entries. On QuALITY, LongHealth, and Long Code Arena, and at every evaluated compaction ratio, \textsc{KV-Surgeon} substantially advances the cost--accuracy Pareto frontier of dense and mixture-of-experts models from 4B to 30B parameters. At $50\times$ compaction, \textsc{KV-Surgeon} improves question-answering accuracy by up to four points over attention matching \citep{zweiger2026fast}, the state of the art, at comparable cost. Lastly, we introduce an even faster variant, which also exceeds the state of the art while requiring only two fifths of its cost.
   \end{abstract}
   
 \section{Introduction}\label{sec:intro}
 Many large language model (LLM) applications repeatedly query a fixed context such as a document, patient record, or codebase.
 The key--value cache (KV-cache) for such a context grows linearly with its length and occupies a large fraction of the memory used during long-context inference. Every attention layer reads this cache for every generated token, so every request pays for its size in both memory and latency. Offline compaction reduces this cost by processing the context once and compressing the resulting cache, which is then used to serve all later requests \citep{eyuboglu2026cartridges,kim2026kvzip,zweiger2026fast}.
 
 Unlike serving-time eviction, which discards cache entries within the forward pass of a request \citep{zhang2023h2o,li2024snapkv}, offline compaction can spend a larger one-time budget that is amortized over future requests. In particular, it can use \emph{calibration queries}, i.e., query vectors the model produces while re-reading or discussing the context, to decide which entries to retain. \citet{eyuboglu2026cartridges} introduce this regime with \emph{Cartridges}, which trains a new latent cache per context. More recent work makes offline compaction much cheaper by replacing end-to-end training with eviction-inspired selection and efficient optimization of the remaining cache \citep{zweiger2026fast}.
 
 The central problem in cache compaction is deciding which cache entries can be removed while preserving the resulting attention outputs. For one attention head, a cache of $T$ entries contains a key $k_j\in\mathbb{R}^{d_k}$ and a value $v_j\in\mathbb{R}^{d_v}$ at each position $j=1,\ldots,T$. A query $q\in\mathbb{R}^{d_k}$ produces
 \begin{equation*}
 P\triangleq\operatorname{softmax}\left(\left(\frac{q^\top k_\ell}{\sqrt{d_k}}\right)_{\ell=1}^{T}\right)\in\mathbb{R}^{T},
 \qquad
 o\triangleq\sum_{j=1}^{T}P_jv_j\in\mathbb{R}^{d_v},
 \end{equation*}
 where $P_j$ is the attention weight of entry $j$ and $o$ is the resulting output. Removing cache entry $j$ does not simply subtract $P_jv_j$, since softmax renormalization changes the attention weights of all remaining entries. The effect of one deletion therefore depends on which other entries remain, making selection a combinatorial problem rather than a ranking problem. Scores based only on received attention, as in attention matching \citep{zweiger2026fast}, additionally overlook value redundancy: if an entry's value is close to the output the head already produces, i.e., $\lVert v_j-o\rVert_2^2$ is small, removing it might change the output little, even when its attention weight is large. Good selection must therefore account not only for the contribution of each entry, but also for how deletions interact.
 
 Recent output-aware methods instead estimate the output error caused by deleting a single entry \citep{goel2026caote,an2026restkvrobustkvcache,gu2025obcache,feng2026criticalkvoptimizingkvcache}. CAOTE \citep{goel2026caote} derives the exact change caused by deleting one entry for a single calibration query, and OBCache \citep{gu2025obcache} uses a Taylor approximation of the attention output error aggregated across queries. These methods nevertheless evaluate the deletion of one entry while the rest of the cache is held fixed. Attention matching \citep{zweiger2026fast} refits the retained cache to minimize the error of the compacted output, but only after the retained set has been chosen. What is missing is \emph{compensation-aware selection}: each candidate deletion should be scored by the error remaining after the surviving entries have been adjusted to compensate for it, so that selection and refitting are optimized \emph{jointly} rather than in sequence. This is difficult under softmax attention because deletion changes the normalization itself, so the error after both deletion and compensation has no obvious closed form.
 
   \begin{figure}[t]
     \centering
     \begin{tikzpicture}
       \begin{axis}[
         intro frontier panel,
         width=0.72\linewidth,
         height=6.6cm,
         xmode=log,
         log ticks with fixed point,
         xmin=2.05, xmax=450,
         ymin=50, ymax=84.5,
         ylabel={QA accuracy (\%)},
       ]
         \addplot[fullctx] coordinates {(0.2,80.72) (1200,80.72)};
         \node[font=\scriptsize, anchor=south west] at (axis cs:2.22,80.72)
           {Full context ($80.7 \pm 0.4\%$)};
   
         \addplot[fold] coordinates {
           (  2.882 ,54.21)
           ( 52.48  ,65.82)
           ( 63.1   ,67.3 )
           (161.7   ,69.49)
           (183.1   ,69.69)
         };
         \addplot[fnew] coordinates {
           (  2.882 ,54.21)
           ( 52.48  ,65.82)
           ( 53.9   ,71.4 )
           (158.5   ,72.53)
         };
   
         \addplot[psum] coordinates {(2.882,54.21)};
         \addplot[pkvz, ebar] table [x=t, y=a] {
           t a
           63.0 53.0
         };
         \addplot[pam2, ebar] table [x=t, y=a] {
           t a
           52.48 65.82
         };
         \addplot[pobc, ebar] table [x=t, y=a] {
           t a
           63.1 67.3
         };
         \addplot[pkvm, ebar] table [x=t, y=a] {
           t a
           82.0 63.9
         };
         \addplot[pobs2, ebar] table [x=t, y=a] {
           t a
           53.9 71.4
         };
         \addplot[pobs, ebar] table [x=t, y=a] {
           t a
           158.5 72.53
         };
         \addplot[pam1, ebar] table [x=t, y=a] {
           t a
           161.7 69.49
         };
         \addplot[pomp1, ebar] table [x=t, y=a] {
           t a
           183.1 69.69
         };
         \addplot[pomps, ebar] table [x=t, y=a] {
           t a
           300.4 69.19
         };
         \node[csum, font=\scriptsize, anchor=north] at (axis cs:3.9,53.3) {Summarization};
         \node[ckvz!85!black, font=\scriptsize, anchor=west, xshift=5pt] at (axis cs:62.0,52.8) {KVzip};
         \node[cobs, font=\scriptsize, anchor=south, xshift=-5pt] at (axis cs:48.9,72.1) {\textsc{KV-Surgeon}-fast};
         \node[cobc!70!black, font=\scriptsize, anchor=north west, xshift=3pt] at (axis cs:62.8,68.1) {OBCache};
         \node[cam!80!black, font=\scriptsize, anchor=north east, xshift=2pt] at (axis cs:48.48,68.9) {AM-HAK-fast};
         \node[ckvm, font=\scriptsize, anchor=north] at (axis cs:82.0,63.2) {KVMerger};
         \node[cobs, font=\scriptsize, anchor=south, yshift=5pt] at (axis cs:158.5,72.0) {\textsc{KV-Surgeon}};
         \draw[cam!85!black, thin] (axis cs:161.7,69.15) -- (axis cs:161.7,63.6);
         \node[cam!85!black, font=\scriptsize, anchor=north] at (axis cs:180,63.4) {AM-HAK};
         \draw[comp!80!black, thin] (axis cs:183.1,69.25) -- (axis cs:200,65.5);
         \node[comp!80!black, font=\scriptsize, anchor=north, align=center] at (axis cs:270,65.9) {AM-OMP-CS};
         \node[comp!70!black, font=\scriptsize, anchor=south, yshift=5pt] at (axis cs:300.0,69.3) {AM-OMP};
       \end{axis}
     \end{tikzpicture}
     \caption{Cost--accuracy frontier at $50\times$ KV-cache compaction on QuALITY with Qwen3-30B-A3B. \textsc{KV-Surgeon} reaches $72.5\%$ accuracy in $159$ seconds and \textsc{KV-Surgeon}-fast reaches $71.4\%$ in $54$ seconds. Pipeline time per context includes calibration-query generation and compaction. OBCache, KVMerger, and KVzip are adapted to our pipeline (\cref{app:abl-regime}). Points are five-seed means. The solid and dash-dotted curves are the Pareto frontiers with and without our methods.}
     \label{fig:frontier-intro}
   \end{figure}
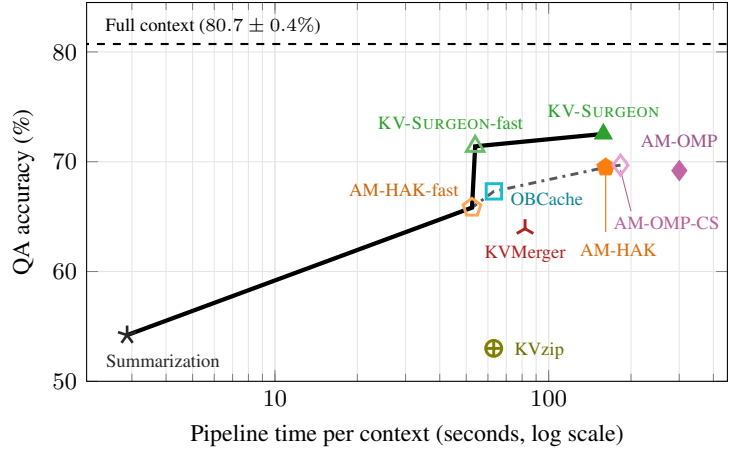
 
 In this work, we substantially advance the cost--accuracy Pareto frontier for offline KV-cache compaction. We derive a quadratic surrogate that makes compensation-aware selection tractable. We introduce \textsc{KV-Surgeon}, which uses this surrogate for greedy, compensation-aware elimination. For each candidate deletion, the surrogate admits closed-form expressions for both the optimal compensating update and the resulting increase in surrogate loss. Each candidate entry is assigned a score after the remaining entries have been adjusted to compensate for its removal, and the update is then applied to the cache, so the next deletion is chosen against the compensated entries, as Optimal Brain Surgeon (OBS) \citep{Hassibi1992} does for network weights.
 
 We evaluate \textsc{KV-Surgeon} on QuALITY, LongHealth, and Long Code Arena, using dense and mixture-of-experts models from 4B to 30B parameters and compaction ratios from $5\times$ to $100\times$. Across these settings, \textsc{KV-Surgeon} improves the cost--accuracy frontier at every ratio, achieving higher mean accuracy at comparable cost. Mean question-answering accuracy across our experiments exceeds the state of the art by $4.3$ percentage points at $100\times$ compaction and $3.0$ percentage points at $50\times$. The gains are less pronounced at $5\times$ and $10\times$, where the strongest methods already suffer almost no deterioration relative to the uncompacted cache. At $50\times$ compaction the state of the art, attention matching with its AM-HAK selection rule \citep{zweiger2026fast}, reaches $62.0\%$ accuracy in $138$ seconds\footnote{Pipeline times differ by more than an order of magnitude between model--benchmark pairs, so an arithmetic mean would be dominated by the slowest pair. We therefore report the geometric mean of the pipeline times.} per context. \textsc{KV-Surgeon} reaches $65.1\%$ in $145$ seconds, and \textsc{KV-Surgeon}-fast reaches $63.9\%$ in $54$ seconds. \Cref{fig:frontier-intro} shows the resulting frontier at $50\times$ compaction on the largest model we evaluate.
 
 Our contributions are the following.
 \begin{enumerate}
   \item \emph{Quadratic surrogate.} We derive a quadratic surrogate to the exact attention-output error, which allows us to design efficient methods for approximately solving the compaction problem.
   \item \emph{Compensation-aware selection.} Based on this surrogate, we introduce \textsc{KV-Surgeon}, which iteratively removes entries while adjusting the remaining entries in closed form after each deletion, then jointly refits the retained cache. Unlike prior output-aware scores or post-selection refitting, compensation directly influences which entries are retained.
   \item \emph{A new cost--accuracy Pareto frontier.} Across QuALITY, LongHealth, and Long Code Arena, dense and mixture-of-experts models from 4B to 30B parameters, and $5\times$ to $100\times$ compaction, \textsc{KV-Surgeon} substantially advances the frontier, with gains at every evaluated ratio and the largest improvements at $50\times$ and $100\times$. We show empirically that accounting for compensation during selection substantially improves accuracy over exact single-deletion scoring.
 \end{enumerate}

   \section{Related work}\label{sec:related}
 
   \textbf{Other forms of KV-cache compression.}
   KV-caches can be compressed along several axes. Quantization lowers the precision of stored keys and values \citep{liu2024kivi,hooper2024kvquant}, and low-rank methods reduce the feature dimension \citep{saxena2024eigenattention,chang2025palu}. Both are complementary to reducing the number of entries. A different route, \emph{LLM summarization}, shortens the text input itself by prompting an LLM to summarize it. This approach transfers across models, but gives no direct control over cache size. We study compaction along the sequence dimension, where a fixed budget of entries is retained from an already computed cache.
   
   \textbf{Offline compaction.}
   Offline methods exploit contexts that are processed once and queried repeatedly. Cartridges \citep{eyuboglu2026cartridges} trains a new latent cache per context from self-generated data. KVzip \citep{kim2026kvzip} removes cache entries based on the attention they receive while the model repeats the context verbatim. Attention matching \citep{zweiger2026fast} also keeps a subset of the original keys, then fits a scalar logit bias per retained key and refits the stored values to reproduce the original attention outputs on reference queries. Its AM-HAK variant selects entries by root-mean-square attention, and AM-OMP, based on orthogonal matching pursuit (OMP), greedily selects keys so that their reweighted attention reproduces the attention mass of the full cache. In both, the support is chosen from attention statistics before the values are refit, so the ability of the retained entries to reproduce the output does not influence selection.
   
   \textbf{Eviction by proxy scores.}
   Serving-time eviction discards entries within a single request, ranking each one by a score that ignores the rest of the cache: accumulated attention, as in H2O \citep{zhang2023h2o} and related methods \citep{oren2024transformers,li2024snapkv,liu2023scissorhands}, or a query-free proxy such as position \citep{xiao2024efficient}, key norm \citep{devoto2024simple}, alignment with a principal query direction \citep{godey2025qfilter}, deviation from neighboring keys \citep{liang2025lagkv}, or value norm \citep{guo2024vatp}. Merging methods soften the same decision by folding evicted entries into retained ones. KVMerger \citep{wang2024kvmerger} merges similar entries into attention-weighted pivots, and \citet{zhang2024cam,wan2024d2o,nawrot2024dmc} use related rules. In all of these, the score of an entry is fixed before any other entry is removed.
   
   \textbf{Output-aware eviction.}
   More recent methods score entries directly through the attention-output error induced by their removal \citep{goel2026caote,an2026restkvrobustkvcache,gu2025obcache,feng2026criticalkvoptimizingkvcache}. CAOTE \citep{goel2026caote} and ReST-KV \citep{an2026restkvrobustkvcache} derive the exact effect of deleting one entry for one query. OBCache \citep{gu2025obcache} derives a second-order approximation in the style of Optimal Brain Damage \citep{LeCun1989}. CriticalKV \citep{feng2026criticalkvoptimizingkvcache} considers deletion of an arbitrary set for one query and derives an additive upper bound on the resulting output error for selection. These methods differ in how they estimate deletion error, but share a common restriction: candidates are evaluated with the retained cache held fixed.
   
   Our work starts from the same output-reconstruction objective as CAOTE, ReST-KV, and OBCache but changes how it is used for selection. Instead of scoring one deletion against a fixed cache, we derive a quadratic surrogate of the output error over arbitrary deletions and reweightings of the retained entries, aggregated over many calibration queries, and let the retained entries compensate during greedy deletion of cache entries.

\section{KV-Surgeon}
\label{sec:prelim}
\label{sec:methods}

In this section, we present our method for offline KV-cache compaction, \textsc{KV-Surgeon}. It compacts each attention head independently by selecting which original cache entries to delete, accounting for how the remaining entries can compensate for deleted ones, and then refits the retained entries to preserve the behavior of the original cache.

\subsection{Problem formulation}
\label{sec:setup}

We consider one layer and one attention head with $T$ cache entries. Cache entry $j$ consists of a key $k_j\in\mathbb{R}^{d_k}$ and a value $v_j\in\mathbb{R}^{d_v}$. We describe the behavior of the head on $N$ \emph{calibration queries} $q_1,\ldots,q_N\in\mathbb{R}^{d_k}$ recorded from the model itself as described in \cref{sec:exp-setup}. For query $i$ and cache entry $j$, the corresponding logit is $Z_{ij}\triangleq q_i^\top k_j/\sqrt{d_k}$. The attention weight and output are
\begin{equation}
    P_{ij}\triangleq
    \frac{e^{Z_{ij}}}{\sum_{\ell=1}^{T}e^{Z_{i\ell}}},
    \qquad
    o_i\triangleq\sum_{j=1}^{T}P_{ij}v_j .
    \label{eq:attention}
\end{equation}
Let $V\in\mathbb{R}^{T\times d_v}$ collect the values as rows and $P\in\mathbb{R}^{N\times T}$ the attention weights. The corresponding output matrix is then $O=PV\in\mathbb{R}^{N\times d_v}$.

\paragraph{Compacted cache.}
Given a budget $t<T$, which is allocated per head, we retain $t$ cache entries corresponding to some index set $S\subseteq\{1,\ldots,T\}$ with $|S|=t$, which we call the \emph{support}. Each retained entry $j\in S$ keeps its original key $k_j$ and is assigned an adapted value $\widehat v_j\in\mathbb{R}^{d_v}$ and a positive attention multiplier $w_j>0$, which rescales the contribution of its entry before the softmax normalization. Together, the remaining keys $k_j$, adapted values $\widehat v_j$, and attention multipliers $w_j$ form the \emph{compacted cache}. For query $i$ and entry $j$, the attention weights and output of the compacted cache are given by
\begin{equation}
  \widehat P_{ij}\triangleq\frac{w_j e^{Z_{ij}}}{\sum_{\ell\in S}w_\ell e^{Z_{i\ell}}} = \frac{w_jP_{ij}}{\sum_{\ell\in S}w_\ell P_{i\ell}} = \frac{w_jP_{ij}}{\mu_i}, \qquad \widehat o_i \triangleq \sum_{j\in S}\widehat P_{ij}\widehat v_j = \frac{1}{\mu_i}\sum_{j\in S}P_{ij}w_j\widehat v_j .
  \label{eq:compactedattention}
\end{equation}
Here,
\begin{equation}
  \mu_i\triangleq\sum_{j\in S}P_{ij}w_j =\frac{\sum_{j\in S}w_j e^{Z_{ij}}} {\sum_{\ell=1}^{T}e^{Z_{i\ell}}}
  \label{eq:mu}
\end{equation}
is the ratio between the compacted and original attention normalization terms. Let $P_S\in\mathbb{R}^{N\times |S|}$ contain the columns of $P$ indexed by $S$, let $w\in\mathbb{R}^{|S|}$ collect the multipliers, and let $\widehat V\in\mathbb{R}^{|S|\times d_v}$ collect the new values as rows. The vector of the normalization ratios is then $\mu\triangleq(\mu_1,\ldots,\mu_N)^\top=P_Sw$ and the output of the compacted cache is $\widehat O=\operatorname{diag}(P_Sw)^{-1}P_S\operatorname{diag}(w)\widehat V$.

\paragraph{Compaction objective.}
Our goal is for the compacted cache to preserve two properties of the original cache. First, its attention outputs $\widehat O$ should be close to the outputs $O$ of the original, uncompacted cache. We measure the output error by
\begin{equation}
    \mathcal{L}(S,w,\widehat V)
    \triangleq
    \lVert \widehat O-O\rVert_F^2 .
    \label{eq:family}
\end{equation}
Second, when combined with other context, such as a question, a system prompt, or other compacted chunks, its entries should collectively receive approximately the same share of attention as the original cache entries, i.e., $\sum_{j\in S}w_j e^{Z_{ij}}\approx\sum_{j=1}^{T}e^{Z_{ij}}$. Output preservation alone does not ensure this: scaling all $w_j$ by the same constant leaves $\widehat O$ unchanged but changes this sum. For a more detailed discussion, see \cref{sec:construction}.

On the calibration queries, we express these two goals as follows, using the ratio $\mu$ of the compacted and original normalization terms defined in \labelcref{eq:mu}
\begin{equation}
    \widehat O\approx O,
    \qquad
    \mu=P_Sw\approx\one .
    \label{eq:requirements}
\end{equation}

\subsection{Tractable surrogate}
\label{sec:surrogate}
Choosing the optimal support $S$ is difficult even if we consider only output preservation. For a fixed support $S$, the output error $\mathcal L(S,w,\widehat V)$ depends on how well the multipliers $w$ and values $\widehat V$ can be adjusted, so evaluating all $\binom{T}{t}$ possible supports is infeasible. In order to approximate this objective, we start from a larger support and sequentially delete entries. However, even assessing the cost of deleting \emph{one} entry $j\in S$ requires re-optimizing $w$ and $\widehat V$ on $S\setminus\{j\}$. Doing that for every candidate deletion at every step is still prohibitively expensive.

We therefore seek a surrogate objective that makes it inexpensive to compare candidate deletions while accounting for how the remaining entries can compensate. After removing an entry, we also want to reuse previous computations to evaluate the next deletions efficiently. Starting from \labelcref{eq:requirements}, we derive a quadratic surrogate with these properties.

\paragraph{Attention as a ratio of two linear terms.}
We introduce the variable $U\triangleq\diag(w)\widehat V\in \mathbb{R}^{|S|\times d_v}$, whose row corresponding to entry $j\in S$ is $u_j^\top=w_j\widehat v_j^\top$ and which absorbs the attention multipliers $w$ into the values $\widehat V$. This is an exact change of variables: since $w_j>0$, we can recover the original values as $\widehat v_j=u_j/w_j$. Substituting into \labelcref{eq:compactedattention}, the output of the compacted cache can be written as the ratio
\begin{equation}
    \widehat o_i=\frac{n_i}{\mu_i},
    \qquad
    n_i\triangleq\sum_{j\in S}P_{ij}u_j\in\mathbb{R}^{d_v},
    \qquad
    \mu_i=\sum_{j\in S}P_{ij}w_j\in\mathbb{R} .
    \label{eq:ratio}
\end{equation}
The numerator $n_i$ and the denominator $\mu_i$ are linear in $U$ and $w$, respectively.

For the original cache, the numerator is $\sum_{j=1}^{T}P_{ij}v_j=o_i$ and the denominator is $\sum_{j=1}^{T}P_{ij}=1$, since every row of $P$ sums to one. Instead of searching for an $\widehat o_i$ matching this ratio, we ask the compacted cache to reproduce numerator and denominator separately, $n_i\approx o_i$ and $\mu_i\approx 1$. The second condition is the mass requirement of \labelcref{eq:requirements}, and together they imply $\widehat o_i\approx o_i$. Define the residuals $r_i\triangleq n_i-o_i$ and $\varepsilon_i\triangleq\mu_i-1$,
\begin{equation}
    \widehat o_i-o_i
    =\frac{n_i-\mu_io_i}{\mu_i}
    =\frac{r_i-\varepsilon_i\,o_i}{\mu_i},
    \qquad\text{hence}\qquad
    \lVert\widehat o_i-o_i\rVert_2\le\frac{\lVert r_i\rVert_2+\lvert\varepsilon_i\rvert\,\lVert o_i\rVert_2}{\mu_i} .
    \label{eq:residuals}
\end{equation}
The output error is thus small whenever both residuals are small: a small $\varepsilon_i$ also keeps $\mu_i$ close to one.

\paragraph{Quadratic surrogate.}
We construct a surrogate of \labelcref{eq:family} by summing the squared residuals from \labelcref{eq:residuals} over all queries. In matrix form, the $r_i$ are the rows of $P_SU-O\in\mathbb{R}^{N\times d_v}$, and the $\varepsilon_i$ are the entries of $\mu-\one=P_Sw-\one\in\mathbb{R}^{N}$. A weight $\gamma^2>0$ controls the relative importance of mass preservation (see \cref{sec:gamma}), giving
\begin{equation}
    \lVert O-P_SU\rVert_F^2
    +
    \gamma^2\lVert\one-P_Sw\rVert_2^2
    =
    \left\lVert
        \begin{bmatrix}O & \gamma\one\end{bmatrix}
        -
        P_S
        \begin{bmatrix}U & \gamma w\end{bmatrix}
    \right\rVert_F^2 .
    \label{eq:decoupled0}
\end{equation}
The original loss \labelcref{eq:family} couples $w$ and $\widehat V$ through their products and the division by $\mu_i$, whereas for a fixed support $S$, the quadratic surrogate separates the optimization of $U$ and $w$ into two linear least-squares problems with the same design matrix $P_S\in\mathbb{R}^{N\times|S|}$. The right-hand side stacks the targets into $[\,O\ \ \gamma\one\,]\in\mathbb{R}^{N\times(d_v+1)}$ and the coefficients into $[\,U\ \ \gamma w\,]\in\mathbb{R}^{|S|\times(d_v+1)}$, yielding a single linear least-squares problem with $d_v+1$ target columns.

\paragraph{Support-selection score.}
In \labelcref{eq:decoupled0}, $U$ is unconstrained, since any $U$ is attained by $\widehat V=\diag(w)^{-1}U$, but the last column of $[\,U\ \ \gamma w\,]$ is constrained by $w>0$. To obtain an unconstrained quadratic problem, we remove this constraint and replace $[\,U\ \ \gamma w\,]$ with a free matrix $B\in\mathbb{R}^{|S|\times(d_v+1)}$. Each row of $B$ corresponds to a retained cache entry $j$: its first $d_v$ coefficients replace the weighted value $w_j\widehat v_j$, and its last coefficient replaces the scaled attention multiplier $\gamma w_j$. These coefficients need not correspond to a valid compacted cache because the last coefficient can be negative. In practice, fewer than half a percent of the retained entries end with a negative coefficient in the last column (\cref{app:abl-gates}), and valid multipliers are restored by the final refit (\cref{sec:refit}).

Finally, we add a small ridge penalty so that the regression has a unique solution for every support. With
$Y\triangleq[\,O\ \ \gamma\one\,]\in\mathbb{R}^{N\times(d_v+1)}$, we define the support score
\begin{equation}
    \mathcal{F}(S)
    \triangleq
    \min_B
    \lVert Y-P_SB\rVert_F^2
    +
    \lambda\lVert B\rVert_F^2,
    \qquad
    \lambda>0 .
    \label{eq:decoupled}
\end{equation}
$Y$ and $\lambda$ stay fixed during support selection: $Y$ depends only on the original cache, and $\lambda$ is set once per head (\cref{sec:app-implementation}). For a fixed support $S$, $\mathcal F(S)$ is the minimum value of this quadratic objective after optimizing all coefficients in $B$. Changing $S$ changes the available columns of $P_S$ and therefore also the optimal coefficients: removing an entry requires the remaining coefficients to compensate for its absence.

\subsection{Greedy elimination with closed-form compensation}
\label{sec:elimination}
\label{sec:algorithm}

With the support score $\mathcal F(S)$ in place, we now develop a greedy elimination algorithm that uses it to choose which cache entries to retain. Starting from a support larger than the target budget $t$, we repeatedly remove the entry whose deletion causes the smallest increase in $\mathcal F(S)$, allowing the remaining coefficients to compensate at each step. The quadratic structure of \labelcref{eq:decoupled} lets us compute both these loss increases and the compensating updates in closed form, using an approach developed in the pruning literature \citep{Hassibi1992,Frantar2023a}. We first describe the optimal coefficients on the current support, then show how to evaluate and apply a deletion.

\paragraph{Optimal coefficients on the current support.}
Each column of $B$ contains the coefficients used to reconstruct one target column of $Y$: either an output coordinate or the attention mass. In \labelcref{eq:decoupled}, these linear least-squares problems all have the same Hessian $2(P_S^\top P_S+\lambda I)$ (see \cref{sec:app-hessian}). A single $|S|\times|S|$ inverse $G$ therefore suffices for all columns. Writing $B$ and $G$ for the optimal coefficients and shared inverse on the current support $S$, we have
\begin{equation}
  B=G\,P_S^\top Y\in\mathbb{R}^{|S|\times(d_v+1)},
  \qquad
  G\triangleq\big(P_S^\top P_S+\lambda I\big)^{-1}\in\mathbb{R}^{|S|\times|S|}.
  \label{eq:ridge-solution}
\end{equation}
The current $B$ and $G$ suffice to evaluate every candidate deletion and obtain the optimal coefficients afterward, without solving another regression problem. For an entry $j\in S$, we define its saliency $s_j$ as the increase in the optimal surrogate objective when that entry is removed:
\begin{equation}
  s_j
  \triangleq
  \mathcal F(S\setminus\{j\})-\mathcal F(S)
  =
  \frac{\lVert B_{j,:}\rVert_2^2}{G_{jj}}.
  \label{eq:saliency}
\end{equation}
The saliency $s_j$ measures the surrogate loss increase \emph{after} the remaining coefficients have been re-optimized. An entry can make a large contribution to the original outputs yet have low saliency if other retained entries can compensate for its removal.

Once an entry $j$ is selected for removal, we obtain the optimal coefficients and inverse for the reduced support directly from the current matrices. Writing $-j$ for the indices in $S\setminus\{j\}$, the updates are
\begin{equation}
    B\leftarrow B_{-j,:}-\frac{G_{-j,j}B_{j,:}}{G_{jj}},
    \qquad
    G\leftarrow G_{-j,-j}-\frac{G_{-j,j}G_{j,-j}}{G_{jj}}.
  \label{eq:obs}
\end{equation}
The updated $B$ is the exact minimizer of \labelcref{eq:decoupled} on $S\setminus\{j\}$, and $G$ is the corresponding inverse Gram matrix. We repeatedly remove the entry with the smallest current saliency, update $B$ and $G$, and recompute the remaining saliencies until $|S|=t$, without solving another regression problem from scratch. A detailed derivation is given in \cref{sec:app-obsproof}.

\subsection{Implementation details}
\label{sec:algdetails}

\textsc{KV-Surgeon} builds on the greedy elimination process described above, with a few practical additions. To reduce computational cost, we restrict the initial set of candidates, delete entries in blocks, and use a cheaper selection rule for heads that retain a relatively large fraction of their original cache entries. After selecting the support, we refit the attention multipliers $w$ and values $\widehat V$ to obtain a valid compacted cache. \Cref{alg:kvsurgeon} summarizes the complete method. We describe these additions below, with further implementation details in \cref{sec:app-implementation} and ablations in \cref{sec:app-abl-design}.

 \begin{algorithm}[h]
   \caption{\textsc{KV-Surgeon} for one attention head}
   \label{alg:kvsurgeon}
   \begin{algorithmic}[1]
     \Require keys $K\in\mathbb{R}^{T\times d_k}$, values $V\in\mathbb{R}^{T\times d_v}$, calibration queries $Q\in\mathbb{R}^{N\times d_k}$, cache budget $t$, pool factor $c$, block size $n_b$, relative ridge $\lambda_0$, mass weight $\gamma_0$, fallback threshold $\tau$, multiplier bound $\beta$
     \algrule
     \State $P\gets\operatorname{softmax}(QK^\top/\sqrt{d_k})$ row-wise, \quad $O\gets PV$
     \State $s^{\mathrm{LOO}}_j\gets\sum_{i=1}^{N}\frac{P_{ij}^2}{(1-P_{ij})^2}\,\lVert v_j-o_i\rVert_2^2$ \quad for $j=1,\ldots,T$
     \If{$t>\tau T$} \Comment{mild-compaction fallback}
       \State $S\gets$ indices of the $t$ largest $s^{\mathrm{LOO}}_j$
       \State \Return $S$, $w=\one$, $\widehat V=V_{S,:}$
     \EndIf
     \State $m\gets\min\{T,\lceil ct\rceil\}$, \quad $S\gets$ indices of the $m$ largest $s^{\mathrm{LOO}}_j$ \Comment{select candidate pool}
     \State $\gamma\gets\gamma_0\operatorname{RMS}_i\lVert o_i\rVert_2$, \quad $Y\gets[\,O\;\;\gamma\one\,]$
     \State $\lambda\gets\lambda_0\,\tr(P_S^\top P_S)/m$
     \State $G\gets(P_S^\top P_S+\lambda I)^{-1}$, \quad $B\gets GP_S^\top Y$
     \While{$|S|>t$} \Comment{greedy elimination}
       \State $s_j\gets\lVert B_{j,:}\rVert_2^2/G_{jj}$ \quad for $j\in S$
       \State $J\gets$ the $\min\{n_b,|S|-t\}$ entries of $S$ with the smallest $s_j$
       \State $R\gets S\setminus J$
       \State $B\gets B_{R,:}-G_{R,J}G_{J,J}^{-1}B_{J,:}$
       \State $G\gets G_{R,R}-G_{R,J}G_{J,J}^{-1}G_{J,R}$
       \State $S\gets R$
     \EndWhile
     \State $w\gets\argmin_{e^{-\beta}\leq w_j\leq e^{\beta}}\ \lVert\one-P_Sw\rVert_2^2$ \Comment{final refit of attention multipliers}
     \State $\widehat P\gets\diag(P_Sw)^{-1}P_S\diag(w)$
     \State $\widehat V\gets\argmin_{\widehat V}\ \lVert\widehat P\widehat V-O\rVert_F^2$ \Comment{final refit of values}
     \algrule
     \State \Return $S$, $w$, $\widehat V$
   \end{algorithmic}
 \end{algorithm}
\textbf{Candidate pool.}
To limit the size of $G$ and the number of deletion steps, we first rank entries by a leave-one-out (LOO) score. We obtain it by removing entry $j$ in \labelcref{eq:compactedattention}, keeping unit multipliers and the remaining values unchanged, and evaluating the output-preservation loss \labelcref{eq:family}:
\begin{equation}
    s_j^{\mathrm{LOO}}
    \triangleq
    \sum_{i=1}^{N}
    \frac{P_{ij}^2}{(1-P_{ij})^2}
    \lVert v_j-o_i\rVert_2^2 .
    \label{eq:loo}
\end{equation}
LOO measures the exact output error on the original cache \emph{without} adjusting the remaining entries, whereas the saliency in \labelcref{eq:saliency} measures the surrogate loss increase accounting for compensation. We keep the $m=\min\{T,\lceil ct\rceil\}$ entries with the largest $s_j^{\mathrm{LOO}}$ and perform greedy elimination only on this pool, using $c=4$. We also use \labelcref{eq:loo} directly as the score of the LOO baseline (\cref{sec:exp-setup}). The score derivation and pool-size ablation are given in \cref{sec:app-loo} and \cref{sec:app-abl-design}, respectively.

\textbf{Block deletion.}
We select a block $J\subset S$ containing the entries with the smallest current saliencies. Analogously to \labelcref{eq:obs}, the optimal coefficients and inverse for the reduced support $R=S\setminus J$ are obtained by
\begin{equation}
    B\leftarrow B_{R,:}-G_{R,J}G_{J,J}^{-1}B_{J,:},
    \qquad
    G\leftarrow G_{R,R}-G_{R,J}G_{J,J}^{-1}G_{J,R},
    \qquad
    R\triangleq S\setminus J.
    \label{eq:block}
\end{equation}
For any selected block $J$, these updates give the exact minimizer of the surrogate on $S\setminus J$. Selecting $J$ by individual saliencies, however, need not minimize the joint deletion cost. Empirically, deleting multiple entries at once improves efficiency without hurting performance. Further implementation details are given in \cref{sec:app-implementation}.

\textbf{Final refit.}\label{sec:refit}
The support-selection procedure determines which entries to retain, but the unconstrained coefficients $B$ need not represent a valid compacted cache. We therefore discard $B$ and $G$ and refit the multipliers $w$ and values $\widehat V$ on the selected support $S$. Following \citet{zweiger2026fast}, we first minimize $\lVert\one-P_Sw\rVert_2^2$ subject to box constraints on $w$, obtaining positive multipliers that preserve the original attention mass as closely as possible. We then fix $w$ and optimize $\widehat V$ to minimize the output-preservation loss \labelcref{eq:family}. The resulting cache retains the original keys with refitted multipliers and values. See \cref{sec:app-implementation} for further details.

\textbf{Mild-compaction fallback.}
For heads that retain a relatively large fraction of their original cache entries, we skip greedy elimination and the final refit, retaining the $t$ entries with the largest LOO scores, their original values, and unit multipliers. Since these heads undergo little compaction, this simpler procedure reduces computation with negligible accuracy loss. Further implementation details are given in \cref{sec:app-implementation} and the corresponding ablation in \cref{sec:app-abl-design}.

\section{Experiments}
 \label{sec:experiments}
We compare \textsc{KV-Surgeon} to existing methods and show that it improves the cost--accuracy frontier across different models, benchmarks, and compaction ratios. We then analyze which part of the algorithm is the source of this improvement through controlled ablations. Finally, we study how much of the gain is retained when self-study calibration is replaced by a single repeat pass. Our code is available at \href{https://github.com/ZIB-IOL/KV-Surgeon}{github.com/ZIB-IOL/KV-Surgeon}.
 
 \subsection{Experimental setup}
 \label{sec:exp-setup} 
 
 \textbf{Benchmarks and models.}
 We evaluate offline KV-cache compaction on long-context multiple-choice question-answering tasks from QuALITY \citep{pang2022qualityquestionansweringlong} and LongHealth \citep{adams2024longhealthquestionansweringbenchmark}, and code completion tasks from Long Code Arena \citep{bogomolov2024long}, where each compacted context is reused across many questions. We evaluate five models, Qwen3-4B, Qwen3-4B-Instruct-2507, Llama-3.1-8B-Instruct, Gemma-3-12B, and Qwen3-30B-A3B \citep{yang2025qwen3,Grattafiori2024,team2025gemma}, spanning three model families, dense and mixture-of-experts architectures, and different attention variants. Qwen3-4B-Instruct-2507 replaces Qwen3-4B on LongHealth and Long Code Arena because it supports longer contexts. We consider $\{100\times, 50\times, 20\times, 10\times, 5\times\}$ compaction, with $50\times$ as the primary operating point. Dataset construction and evaluation details are given in \cref{sec:expdetails}.
 
 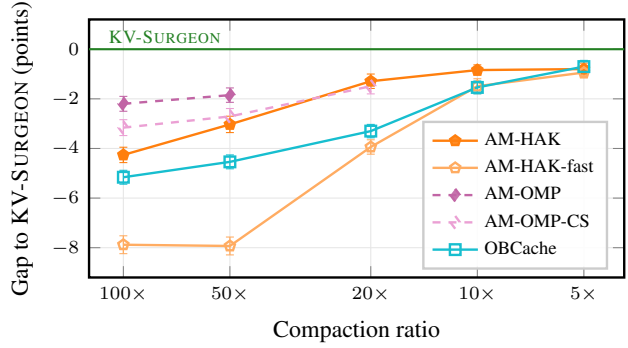
\begin{figure}[t]
   \centering
   \begin{tikzpicture}
     \begin{axis}[
       ratio panel,
       xmode=log,
       log ticks with fixed point,
       width=0.62\linewidth, height=5cm,
       ylabel={Gap to \textsc{KV-Surgeon} (points)},
       ymin=-9.2, ymax=1.2,
       legend pos=south east,
       legend columns=1,
       error bars/y dir=both, error bars/y explicit,
       error bars/error bar style={line width=0.5pt},
       error bars/error mark options={rotate=90, mark size=1.5pt},
     ]
       \addplot[cobs!80!black, thick, forget plot, mark=none]
         coordinates {(0.008,0) (0.26,0)};
       \node[font=\scriptsize, color=cobs!80!black, anchor=south west] at (axis cs:0.0085,-0.15) {\textsc{KV-Surgeon}};
       \addplot[lam] coordinates {(0.01,-4.26) +- (0,0.31) (0.02,-3.03) +- (0,0.33) (0.05,-1.29) +- (0,0.29) (0.1,-0.84) +- (0,0.21) (0.2,-0.80) +- (0,0.24)};
       \addlegendentry{AM-HAK}
       \addplot[lamf] coordinates {(0.01,-7.88) +- (0,0.36) (0.02,-7.93) +- (0,0.36) (0.05,-3.94) +- (0,0.29) (0.1,-1.49) +- (0,0.30) (0.2,-0.94) +- (0,0.22)};
       \addlegendentry{AM-HAK-fast}
       \addplot[lomp] coordinates {(0.01,-2.20) +- (0,0.30) (0.02,-1.85) +- (0,0.29)};
       \addlegendentry{AM-OMP}
       \addplot[lompf] coordinates {(0.01,-3.16) +- (0,0.32) (0.02,-2.71) +- (0,0.32) (0.05,-1.49) +- (0,0.31)};
       \addlegendentry{AM-OMP-CS}
       \addplot[lobc] coordinates {(0.01,-5.16) +- (0,0.28) (0.02,-4.54) +- (0,0.28) (0.05,-3.30) +- (0,0.27) (0.1,-1.54) +- (0,0.26) (0.2,-0.70) +- (0,0.25)};
       \addlegendentry{OBCache}
     \end{axis}
   \end{tikzpicture}
   \caption{Mean accuracy over the eight pairs of \cref{tab:main-acc} relative to \textsc{KV-Surgeon}. Error bars are one standard error of the difference. AM-OMP and AM-OMP-CS are shown only at the ratios their solver reaches on every model and benchmark.}
   \label{fig:mean-ratio}
 \end{figure}
 \textbf{Protocol.}
 All accuracies are means over five random seeds, with one standard error reported in the tables. All methods shown in the main comparison, except LLM summarization, use heterogeneous allocation of the global cache budget across heads. The allocation is computed once per model on data disjoint from the evaluation benchmarks following \citet{zweiger2026fast} and kept fixed between methods. All methods using calibration additionally share the attention-multiplier parameterization and final cache refit of \cref{sec:prelim}, so their comparisons isolate differences in support selection. \textsc{KV-Surgeon} hyperparameters are fixed once on Qwen3-4B with QuALITY at $50\times$ compaction and reused for all other models, benchmarks, and compaction ratios, i.e., \textsc{KV-Surgeon} does not require per-model or per-benchmark tuning.

 \textbf{Calibration and cost.} To collect calibration queries, we use two different approaches. Cheap calibration queries are obtained by a \emph{repeat pass}, which concatenates the context with a ``repeat verbatim'' prompt and another copy of the context \citep{kim2026kvzip}. For \emph{self-study}, the more expensive approach, we prompt the model to discuss the context multiple times and record the queries produced during answer generation \citep{eyuboglu2026cartridges}. The suffix \emph{-fast} denotes repeat-pass calibration for \textsc{KV-Surgeon} and AM-HAK. AM-OMP-CS is the cheap-solver setting of \citet{zweiger2026fast}, named AM-OMP-fast in the original work, and uses self-study queries. Pipeline time covers calibration-query generation and compaction, but excludes the context prefill shared by all methods and preprocessing performed once per model. Further timing and calibration details are given in \cref{sec:expdetails}.
 
 \textbf{Baselines.}
 Our main prior baseline is attention matching \citep{zweiger2026fast}, for which we evaluate both AM-HighestAttentionKeys (AM-HAK) and AM-OMP using the authors' implementation and hyperparameters. We also adapt several serving-time eviction methods to the offline setting (\cref{app:abl-regime}) and include our adaptation of OBCache in the main experiments. LOO ranks entries by the exact single-deletion score of \labelcref{eq:loo} and shares \textsc{KV-Surgeon}'s calibration queries, head-budget allocation, and final refit, providing a controlled comparison to independent selection. We additionally include LLM summarization and query-free baselines (\cref{sec:app-grids}).

 \begin{table}[t]
   \centering
   \caption{
     Accuracy (\%) at $50\times$ compaction. Entries are means over five seeds with standard error over seeds in \textcolor{black!55}{grey}. Within each group and column, bold denotes the highest mean and underlining denotes results that trail the highest mean by at most two standard errors of the difference. The Qwen3-4B column uses Qwen3-4B on QuALITY and Qwen3-4B-Instruct-2507 on LongHealth. Summarization uses the prompt whose summary length is closest to $50\times$ compaction (\cref{sec:app-grids}). The \emph{Mean} column contains the means over the eight model--benchmark pairs, and \emph{Time} is the geometric mean of pipeline time per context in seconds.}
  \vspace{9pt}
   \footnotesize
   \setlength{\tabcolsep}{2pt}
   \resizebox{\linewidth}{!}{
   \begin{tabular}{@{}lcccccccccc@{\hskip 12pt}c@{}}
     \toprule
     \multirow{2}{*}{Method} & \multirow{2}{*}{Time (s)} & \multicolumn{2}{c}{Qwen3-4B}
     & \multicolumn{2}{c}{Llama-3.1-8B}
     & \multicolumn{2}{c}{Gemma-3-12B}
     & \multicolumn{2}{c}{Qwen3-30B} & & \multirow{2}{*}{Mean} \\
     \cmidrule(lr){3-4} \cmidrule(lr){5-6} \cmidrule(lr){7-8} \cmidrule(lr){9-10}
     & & QuAL. & LongH. & QuAL. & LongH. & QuAL. & LongH. & QuAL. & LongH. & & \\
     \midrule
     \multicolumn{11}{@{}l}{\emph{Repeat-pass}} \\
     \textsc{KV-Surgeon}-fast (ours) & 54 & \textbf{65.9}\se{0.4} & \textbf{66.9}\se{0.5} & \textbf{61.0}\se{0.4} & \textbf{61.5}\se{0.6} & \textbf{68.1}\se{0.5} & \textbf{52.1}\se{0.4} & \textbf{71.4}\se{0.3} & \textbf{64.1}\se{0.2} & & \textbf{63.9}\se{0.15} \\
     OBCache (adapted) & 58 & 62.4\se{0.7} & 62.8\se{0.9} & 59.8\se{0.3} & 52.4\se{0.2} & 66.6\se{0.4} & 50.2\se{0.8} & 67.3\se{0.4} & 62.8\se{0.4} & & 60.5\se{0.20} \\
     AM-HAK-fast & 54 & 58.1\se{0.5} & 53.5\se{1.3} & 58.7\se{0.6} & 53.5\se{1.1} & 62.2\se{0.4} & 45.6\se{1.2} & 65.8\se{0.5} & 59.8\se{0.6} & & 57.1\se{0.30} \\
     \midrule
     \multicolumn{11}{@{}l}{\emph{Self-study}} \\
     \textsc{KV-Surgeon} (ours) & 145 & \textbf{66.8}\se{0.3} & \textbf{67.8}\se{0.8} & \textbf{62.3}\se{0.5} & \textbf{62.1}\se{0.3} & \underline{67.5}\se{0.7} & \textbf{56.4}\se{0.7} & \textbf{72.5}\se{0.6} & \textbf{65.2}\se{0.4} & & \textbf{65.1}\se{0.20} \\
     LOO & 139 & 65.6\se{0.2} & 64.2\se{1.2} & \underline{61.2}\se{0.7} & 59.1\se{0.6} & 65.1\se{0.5} & 49.6\se{0.7} & 69.6\se{0.3} & 62.3\se{0.7} & & 62.1\se{0.24} \\
     AM-HAK & 138 & 64.1\se{0.9} & 65.0\se{0.8} & 60.4\se{0.5} & 58.2\se{0.5} & 65.1\se{0.3} & 52.7\se{1.4} & 69.5\se{0.3} & 61.4\se{0.4} & & 62.0\se{0.26} \\
     AM-OMP-CS & 266 & 65.7\se{0.4} & \underline{65.9}\se{1.1} & \underline{60.6}\se{0.8} & 55.8\se{0.9} & \underline{67.2}\se{0.2} & 50.9\se{0.8} & 69.7\se{0.5} & 63.1\se{0.2} & & 62.4\se{0.24} \\
     AM-OMP & 763 & \underline{66.4}\se{0.2} & 65.5\se{0.6} & \underline{61.3}\se{0.6} & 57.3\se{0.9} & \textbf{67.7}\se{0.3} & \underline{54.5}\se{0.9} & 69.2\se{0.4} & \underline{63.9}\se{0.6} & & 63.2\se{0.22} \\
     \midrule
     Q-Filters (query-free) & 1.6 & 52.4\se{0.5} & 43.6\se{0.4} & 47.7\se{0.8} & 36.1\se{1.1} & 55.9\se{0.3} & 39.0\se{0.7} & 50.9\se{0.3} & 43.2\se{0.5} & & 46.1\se{0.22} \\
     Summarization & 5.1 & 51.2\se{0.5} & 28.0\se{0.7} & 46.7\se{0.3} & 31.5\se{0.7} & 50.2\se{0.4} & 39.4\se{0.4} & 54.2\se{0.6} & 47.2\se{0.6} & & 43.6\se{0.19} \\
     Full context & -- & 72.1\se{0.3} & 80.5\se{0.8} & 62.7\se{0.5} & 73.0\se{0.8} & 70.6\se{0.4} & 74.5\se{0.4} & 80.7\se{0.4} & 78.3\se{0.5} & & 74.0\se{0.19} \\
     \bottomrule
   \end{tabular}}
   \label{tab:main-acc}
 \end{table}
 
 \subsection{KV-Surgeon improves the cost--accuracy frontier}
 \label{sec:exp-main}

 At $50\times$ compaction, both \textsc{KV-Surgeon} and \textsc{KV-Surgeon}-fast exceed the prior state of the art (\cref{tab:main-acc}). \textsc{KV-Surgeon}-fast reaches $63.9\%$ mean accuracy in $54$ seconds per context, compared with $63.2\%$ for AM-OMP in $763$ seconds, a $14.1\times$ reduction in pipeline time. Self-study calibration improves \textsc{KV-Surgeon} further to $65.1\%$ in $145$ seconds, $1.9$ percentage points above AM-OMP at $5.3\times$ lower time. At comparable cost, \textsc{KV-Surgeon}-fast is $3.3$ percentage points ahead of adapted OBCache, while \textsc{KV-Surgeon} is $3.0$ points ahead of AM-HAK.
 
 These gains are consistent across models and benchmarks. For both calibration methods, \textsc{KV-Surgeon} is best or tied for best within uncertainty on all eight model--benchmark pairs, while \textsc{KV-Surgeon}-fast is ahead of every other repeat-pass method beyond uncertainty. Thus, the more expensive AM-OMP variants no longer lie on the cost--accuracy frontier at our primary operating point (\cref{fig:frontier-intro}).

 The advantage of \textsc{KV-Surgeon} also persists across compaction ratios. \Cref{fig:mean-ratio} shows mean accuracy over the eight model--benchmark pairs relative to \textsc{KV-Surgeon}. All prior methods are worse by more than two standard errors at every compaction ratio. The gap is largest between $100\times$ and $20\times$ compaction and narrows at $10\times$ and $5\times$, where all methods approach the performance of the full-context ceiling. We therefore focus the analysis below on large compaction ratios, where support selection has the largest effect.
 
 The same compute-time advantage transfers to project-level code completion on Long Code Arena (\cref{tab:lca-qwen4b}, \cref{sec:app-lca}). At $50\times$, \textsc{KV-Surgeon} and \textsc{KV-Surgeon}-fast match AM-OMP within uncertainty while using $9.7$ and $21.5$ times less compute time, respectively. At $20\times$, \textsc{KV-Surgeon} has the highest point estimate, with AM-OMP-CS the only prior method within uncertainty at $6.7$ times higher cost.
 
 \begin{table}[t]
   \centering
   \caption{Qwen3-4B on QuALITY (first $25$ articles) at $50\times$ compaction. We compare the impact of compensation during selection. We vary the selection rule, i.e., the saliency by which entries are ranked for deletion, and whether the surviving entries are updated after each deletion.}
   \vspace{9pt}
   \small
   \setlength{\tabcolsep}{7pt}
   \begin{tabular}{@{}llcc@{}}
       \toprule
       Method & Selection rule & Compensation & Accuracy (\%)  \\
       \midrule
       \textsc{KV-Surgeon} & \cref{eq:saliency} & \cmark & \textbf{66.9} \\
       \textsc{KV-Surgeon} w/o compensation & \cref{eq:saliency} & \xmark & 63.9  \\
       LOO & \cref{eq:loo} & \xmark & 64.7 \\
       \bottomrule
   \end{tabular}
   \label{tab:comp-ablation}
 \end{table}

 \subsection{Sources of the performance gain}
 \label{sec:exp-obs}
 The gain over independent selection comes from compensating the surviving cache entries as the support changes. To isolate this effect, we keep the surrogate, candidate pool, calibration queries, and final refit fixed and disable only the compensating update: we delete entries based on the saliency \labelcref{eq:saliency} but do not apply the update \labelcref{eq:obs} to the remaining entries, so the saliencies stay fixed and the elimination reduces to a single ranking of the candidate pool. At $50\times$ compaction on Qwen3-4B with QuALITY, this reduces accuracy from $66.9\%$ to $63.9\%$ (\cref{tab:comp-ablation}). \textsc{KV-Surgeon} without compensation also performs similarly to LOO, which uses the different deletion rule \labelcref{eq:loo} and does not adapt the remaining entries at all. Replacing the saliency score alone does not explain the improvement. Rather, the gain comes from updating the remaining entries \emph{during} selection. Additional controls that remove the opportunity for iterative compensation lead to the same conclusion (\cref{sec:app-ablations,sec:app-surrogate}).
 
 The improvement is also not an artifact of the optimized allocation of the global cache budget across heads inherited from \citet{zweiger2026fast}. Under uniform allocation, \textsc{KV-Surgeon} remains $3.0$ accuracy points ahead of AM-HAK (\cref{sec:app-ablations}). Compensation-aware support selection therefore provides an improvement that is orthogonal to that of the offline-compaction pipeline proposed by \citet{zweiger2026fast}.
 
 \subsection{Cheap calibration preserves most of the gain}
 \label{sec:exp-fast}
 
 \textsc{KV-Surgeon} is comparatively insensitive to how calibration queries are collected. Replacing self-study with a single repeat pass reduces pipeline time from $145$ to $54$ seconds per context at $50\times$ compaction while reducing mean accuracy by only $1.2$ points. The same change is substantially more costly for AM-HAK, which loses $4.9$ points (\cref{tab:main-acc}). This makes \textsc{KV-Surgeon}-fast not only cheaper, but sufficient to outperform all prior methods at the primary operating point.
 
 This robustness persists across compaction ratios. Replacing self-study with repeat-pass calibration costs \textsc{KV-Surgeon} at most $1.2$ points of mean accuracy at any compaction ratio, with the gap shrinking as more cache entries are retained. Detailed results across ratios are reported in \cref{sec:app-grids}; \cref{sec:app-surrogate} offers a hypothesis for this robustness.
 
 \section{Conclusion}
 \label{sec:conclusion}
 Prior offline KV-cache compaction methods first deleted cache entries and then adapted the remaining ones. We showed that deletion and adaptation can be coupled efficiently, and that doing so improves accuracy. We derived a quadratic surrogate of the attention-output error that allows for an efficient greedy algorithm with compensation-aware selection: every candidate deletion is scored after the retained entries have compensated for it. The resulting method, \textsc{KV-Surgeon}, improves the cost--accuracy frontier at every compaction ratio on QuALITY, LongHealth, and Long Code Arena with models from 4B to 30B parameters, and its fast variant exceeds all prior methods at two fifths of the cost of the strongest baseline. A controlled comparison attributes the gain over independent selection to compensation during selection, not to the surrogate.
 
 \section*{Acknowledgment}
 \label{sec:app-acknowledgments}
 This research was partially supported by the DFG Cluster of Excellence MATH+ (EXC-2046/2, project id 390685689) funded by the Deutsche Forschungsgemeinschaft (DFG) as well as by the German Federal Ministry of Research, Technology and Space (research campus Modal, fund number 05M14ZAM, 05M20ZBM) and the VDI/VDE Innovation + Technik GmbH (fund number 16IS23025B).
 
 We would also like to thank the Speinshart Scientific Center for AI and SuperTech, where part of this work was conducted, for providing such a stimulating research environment.

 \bibliography{iclr2026_conference}
 \bibliographystyle{iclr2026_conference}
 
 \appendix
\crefalias{section}{appendix}
\crefalias{subsection}{subappendix}
\crefalias{subsubsection}{subsubappendix}
 \newpage

 \section{Derivations}
 \label{sec:app-derivations}
 
 \subsection{The surrogate as a second-order Taylor expansion}
 \label{sec:app-taylor}

 An alternative route to the same surrogate starts with the exact output loss after substituting $U=\diag(w)\widehat V$:
 \begin{equation}
   \mathcal L(S,w,\widehat V)
   =\sum_{i=1}^{N}\frac{1}{\mu_i^2}
   \left\lVert\left(\sum_{j\in S}P_{ij}u_j-o_i\right)+(1-\mu_i)o_i\right\rVert_2^2.
   \label{eq:compactionproblem}
 \end{equation}
 Replacing $\mu_i^{-2}$ by its value of one at the original cache gives the quadratic
 \begin{equation}
   \mathcal Q(S,w,U)\triangleq
   \sum_{i=1}^{N}\left\lVert\sum_{j\in S}P_{ij}u_j-\mu_i o_i\right\rVert_2^2.
   \label{eq:frozen}
 \end{equation}
 Dropping the cross term between the output and mass residuals separates the two terms:
 \begin{equation}
   \lVert O-P_SU\rVert_F^2
   +\sum_{i=1}^{N}\bigl(\lVert o_i\rVert_2(1-(P_Sw)_i)\bigr)^2.
   \label{eq:split}
 \end{equation}
 Replacing the query-dependent weights $\lVert o_i\rVert_2$ by the shared $\gamma$ then gives \labelcref{eq:decoupled0}.
 The quadratic $\mathcal Q$ in \labelcref{eq:frozen} is the second-order Taylor expansion of $\mathcal L$ around the original cache. This supplies an alternative derivation of the surrogate in \cref{sec:surrogate}. The method represents a deletion by removing an index from $S$, and a support $S$ and the original cache have parameters of different sizes. To place them in a common coordinate system we use, in this proof only, a full-length multiplier vector $m\in\mathbb{R}^T$ with $m_j=w_j$ for $j\in S$ and $m_j=0$ otherwise, and $\widetilde U\in\mathbb{R}^{T\times d_v}$ with rows $m_j\widehat v_j$, which vanish outside $S$. Then $\sum_{j\in S}P_{ij}u_j^\top=P_i\widetilde U$ and $\mu_i=P_im$, where $P_i$ is row $i$ of $P$, so that
 \begin{equation*}
     \mathcal{L}=\sum_{i=1}^{N}\frac{\bigl\lVert P_i\widetilde U-(P_im)\,o_i\bigr\rVert_2^2}{(P_im)^2},
     \qquad
     \mathcal{Q}=\sum_{i=1}^{N}\bigl\lVert P_i\widetilde U-(P_im)\,o_i\bigr\rVert_2^2
 \end{equation*}
 are functions of $(\widetilde U,m)\in\mathbb{R}^{T\times d_v}\times\mathbb{R}^T$ whose domain contains the original cache $(\widetilde U,m)=(V,\one)$. There the vector inside the norms is $P_iV-(P_i\one)\,o_i=o_i-o_i=0$ for every query, because $P_iV=o_i$ by the definition of $O$ and $P_i\one=1$ since every row of $P$ is a softmax; hence $\mathcal{L}$ and $\mathcal{Q}$ vanish together with their gradients. Write $(\widetilde U,m)=(V+\Delta_U,\one+\Delta_m)$. Substituting, the vector inside the norms becomes $P_i(V+\Delta_U)-\bigl(P_i(\one+\Delta_m)\bigr)\,o_i=(P_iV-o_i)+P_i\Delta_U-(P_i\Delta_m)\,o_i=P_i\Delta_U-(P_i\Delta_m)\,o_i$, again by $P_iV=o_i$ and $P_i\one=1$. This is linear in $(\Delta_U,\Delta_m)$, so its squared norm is a homogeneous quadratic in the perturbation, while $(P_im)^{-2}=1-2P_i\Delta_m+O(\lVert\Delta_m\rVert_2^2)$. Hence $\mathcal{L}=\mathcal{Q}+O(\lVert(\Delta_U,\Delta_m)\rVert^3)$, and $\mathcal{Q}$, itself a quadratic, is the second-order Taylor polynomial of $\mathcal{L}$ at the original cache. As in OBS \citep{Hassibi1992}, the expansion is then used at deletions, which move $m_j$ from one to zero and row $j$ of $\widetilde U$ from $v_j$ to zero and are not small perturbations.
 
 For unchanged values, $\widetilde U=\diag(m)V$, the two terms inside the norms are $P_i\diag(m)V=\sum_jm_jP_{ij}v_j$ and $(P_im)\,o_i=\sum_jm_jP_{ij}\,o_i$, so the vector inside the norms is $\sum_jm_jP_{ij}(v_j-o_i)$ and the frozen quadratic is the quadratic form
 \begin{equation*}
     \mathcal{Q}=m^\top Hm,
     \qquad
     H_{jk}\triangleq\sum_{i=1}^{N}P_{ij}P_{ik}\,(v_j-o_i)^\top(v_k-o_i).
 \end{equation*}
 Its off-diagonal entries record whether the errors of deleting $j$ and $k$ reinforce or cancel. Its diagonal $H_{jj}=\sum_iP_{ij}^2\lVert v_j-o_i\rVert_2^2$ is the second-order truncation of the exact single-deletion loss, the score of \labelcref{eq:loo} without the factor $(1-P_{ij})^{-2}$ (\cref{sec:app-loo}). Finally, $H\one=0$ because $\sum_kP_{ik}(v_k-o_i)=0$. This is the null direction of the Hessian along uniform scalings of all weights, which leave the ratio of \labelcref{eq:compactedattention} unchanged; the frozen quadratic alone therefore does not determine the scale of the multipliers, and the mass term of \labelcref{eq:decoupled0} does (\cref{sec:app-rms,sec:mass}).

 \subsection{Hessian structure of the surrogate}
 \label{sec:app-hessian}

 We compare the quadratic objectives of \cref{sec:surrogate} through their Hessians in the unknowns $(U,w)$. Stack the unknowns as $x\triangleq(U_{:,1},\ldots,U_{:,d_v},w)\in\mathbb{R}^{|S|(d_v+1)}$, where $U_{:,c}$ is column $c$ of $U$, and write $O_{:,c}$ for column $c$ of $O$.

 \textbf{After freezing the normalization.} The frozen objective \labelcref{eq:frozen} is $\lVert P_SU-\diag(P_Sw)O\rVert_F^2$. Column $c$ of $\diag(P_Sw)O$ is $\diag(O_{:,c})P_Sw$, so
 \begin{equation*}
     \sum_{i=1}^{N}\Big\lVert\Big(\sum_{j\in S}P_{ij}u_j-o_i\Big)+(1-\mu_i)o_i\Big\rVert_2^2
     =\sum_{c=1}^{d_v}\big\lVert P_SU_{:,c}-\diag(O_{:,c})P_Sw\big\rVert_2^2 .
 \end{equation*}
 This is a quadratic in $x$. Its Hessian has the diagonal blocks $2P_S^\top P_S$ for each column $U_{:,c}$, the diagonal block $2P_S^\top\diag(\lVert o_1\rVert_2^2,\ldots,\lVert o_N\rVert_2^2)P_S$ for $w$, and the off-diagonal blocks $-2P_S^\top\diag(O_{:,c})P_S$ between $U_{:,c}$ and $w$. The off-diagonal blocks are the Hessian of the cross term in \labelcref{eq:frozen}: expanding the square gives $2\sum_i\langle\sum_jP_{ij}u_j-o_i,\,(1-\mu_i)o_i\rangle$, whose bilinear part in $(U,w)$ is $-2\sum_cU_{:,c}^\top P_S^\top\diag(O_{:,c})P_Sw$. Because of these blocks, the columns of $U$ and the multipliers are not separate regressions on $P_S$. Scoring a deletion, or computing the coefficients after one, requires the inverse of the full Hessian of dimension $|S|(d_v+1)$ instead of an $|S|\times|S|$ matrix, which multiplies the memory by $(d_v+1)^2$ and the cost of every update by at least the same factor.

 \textbf{After dropping the cross term.} The objective \labelcref{eq:split} is a function of $U$ plus a function of $w$, so its Hessian is block diagonal, with the blocks $2P_S^\top P_S$ for the columns of $U$ and the block $2P_S^\top\diag(\lVert o_1\rVert_2^2,\ldots,\lVert o_N\rVert_2^2)P_S$ for $w$. Every column can be optimized on its own. The multipliers, however, have a different Gram matrix from the values, so greedy elimination would carry two inverses and update both after every deletion.

 \textbf{After replacing the weights by $\gamma$.} In \labelcref{eq:decoupled0} the mass term is $\gamma^2\lVert\one-P_Sw\rVert_2^2=\lVert\gamma\one-P_S(\gamma w)\rVert_2^2$, a regression of the column $\gamma\one$ on $P_S$ in the unknown $\gamma w$, with Gram matrix $P_S^\top P_S$. All $d_v+1$ diagonal blocks of the Hessian are now $2P_S^\top P_S$ and there are no off-diagonal blocks. The ridge in \labelcref{eq:decoupled} adds $2\lambda I$ to each block, so the Hessian of \labelcref{eq:decoupled} in $B$ consists of $d_v+1$ identical blocks $2(P_S^\top P_S+\lambda I)$.

 \textbf{Closed form.} Expanding \labelcref{eq:decoupled} gives
 \begin{equation*}
     \lVert Y-P_SB\rVert_F^2+\lambda\lVert B\rVert_F^2
     =\tr(Y^\top Y)-2\tr(B^\top P_S^\top Y)+\tr\big(B^\top(P_S^\top P_S+\lambda I)B\big).
 \end{equation*}
 The matrix $P_S^\top P_S+\lambda I$ is positive definite for $\lambda>0$, since $x^\top(P_S^\top P_S+\lambda I)x=\lVert P_Sx\rVert_2^2+\lambda\lVert x\rVert_2^2>0$ for $x\neq0$. The objective is therefore strictly convex in $B$, and its unique minimizer is the stationary point $(P_S^\top P_S+\lambda I)B=P_S^\top Y$, that is, $B=GP_S^\top Y$ with $G=(P_S^\top P_S+\lambda I)^{-1}$, which is \labelcref{eq:ridge-solution}. Column $c$ of $B$ is $GP_S^\top Y_{:,c}$, so one inverse serves all columns. Inserting the minimizer and using $(P_S^\top P_S+\lambda I)B=P_S^\top Y$ for the first equality and $G(P_S^\top P_S+\lambda I)G=G$ for the second,
 \begin{equation}
     \mathcal F(S)
     =\tr(Y^\top Y)-\tr\big(B^\top(P_S^\top P_S+\lambda I)B\big)
     =\tr(Y^\top Y)-\tr\big(Y^\top P_SGP_S^\top Y\big).
     \label{eq:Fvalue}
 \end{equation}

 \subsection{Scale of the mass term}
 \label{sec:app-rms}
 \label{sec:gamma}

 In \labelcref{eq:split} the mass error of query $i$ is weighted by $\lVert o_i\rVert_2^2$: a query with a large output is hurt more by a given relative mass deficit. These weights depend neither on $w$ nor on $S$, so the mass term of \labelcref{eq:split} could be minimized as it stands, with the Gram matrix $P_S^\top\diag(\lVert o_1\rVert_2^2,\ldots,\lVert o_N\rVert_2^2)P_S$. Carrying it through the elimination would require a second inverse next to $G$, updated after every deletion. Replacing $\lVert o_i\rVert_2$ by one common scalar $\gamma$ gives the mass term the Gram matrix $P_S^\top P_S$ of the output term, so that one inverse serves all $d_v+1$ columns (\cref{sec:app-hessian}).

 We set $\gamma\triangleq\gamma_0\operatorname{RMS}_i\lVert o_i\rVert_2=\gamma_0\,(N^{-1}\sum_i\lVert o_i\rVert_2^2)^{1/2}$ with a dimensionless $\gamma_0>0$. The RMS keeps the mass column of $Y$ on the scale of the output columns: for a uniform relative mass deficit $1-\mu_i\equiv\epsilon$, the mass term of \labelcref{eq:split} is $\epsilon^2\sum_i\lVert o_i\rVert_2^2$ and that of \labelcref{eq:decoupled0} is $N\gamma^2\epsilon^2$, and the two agree for every $\epsilon$ if and only if $\gamma_0=1$. The factor $\gamma_0$ therefore sets the weight of the mass term relative to the output term, with $\gamma_0=1$ matching the cost that a uniform deficit has under \labelcref{eq:split}. We use $\gamma_0=1$ throughout (\cref{sec:expdetails}). Setting $\gamma_0=0$ removes the mass term from the surrogate and changes accuracy within one standard error (\cref{sec:app-abl-design}).

 \subsection{Deletion saliencies and compensation}
 \label{sec:app-obsproof}

 We derive the saliency \labelcref{eq:saliency} and the updates \labelcref{eq:obs,eq:block} from the block structure of the inverse $G$. Throughout, $S$ is the current support, $A\triangleq P_S^\top P_S+\lambda I$, $G=A^{-1}$, and $B=GP_S^\top Y$ as in \labelcref{eq:ridge-solution}. Rows and columns of $A$, $G$, and $B$ are indexed by the elements of $S$, so that an entry keeps its index when other entries are deleted. For $j\in S$, $-j$ denotes the index set $S\setminus\{j\}$: $G_{-j,-j}$ is $G$ without row and column $j$, $G_{-j,j}$ is column $j$ of $G$ without its $j$-th entry, and $B_{-j,:}$ is $B$ without row $j$. We write $p_j\in\mathbb{R}^N$ for column $j$ of $P$, the attention weights of all queries on entry $j$.

 \textbf{Deleting an entry removes a row and a column of $A$.} Entry $(k,l)$ of $P_S^\top P_S$ is $p_k^\top p_l$ and involves only the columns $k$ and $l$. Removing column $j$ from $P_S$ removes row and column $j$ from $P_S^\top P_S$ and changes no other entry, and $\lambda I$ loses the same row and column. The Gram matrix of the reduced support is therefore the principal submatrix
 \begin{equation}
     P_{S\setminus\{j\}}^\top P_{S\setminus\{j\}}+\lambda I=A_{-j,-j} .
     \label{eq:principal}
 \end{equation}
 This requires the same $\lambda$ on both supports: a ridge that changes with $S$ would change the remaining entries as well.

 \textbf{Block inverse.} The block equations of $AG=I$ for the partition $(\{j\},-j)$ are
 \begin{equation*}
     A_{jj}G_{jj}+A_{j,-j}G_{-j,j}=1,\qquad
     A_{-j,j}G_{jj}+A_{-j,-j}G_{-j,j}=0,\qquad
     A_{-j,j}G_{j,-j}+A_{-j,-j}G_{-j,-j}=I .
 \end{equation*}
 $A_{-j,-j}$ is invertible as a principal submatrix of a positive definite matrix, and $G_{jj}=e_j^\top A^{-1}e_j>0$. The second equation gives $A_{-j,j}=-A_{-j,-j}G_{-j,j}/G_{jj}$, and substituting this into the third gives $A_{-j,-j}\big(G_{-j,-j}-G_{-j,j}G_{j,-j}/G_{jj}\big)=I$. Hence
 \begin{equation}
     (A_{-j,-j})^{-1}=G_{-j,-j}-\frac{G_{-j,j}G_{j,-j}}{G_{jj}},
     \qquad
     \frac{1}{G_{jj}}=A_{jj}-A_{j,-j}(A_{-j,-j})^{-1}A_{-j,j},
     \label{eq:blockinverse}
 \end{equation}
 where the second identity follows from the first block equation after inserting $(A_{-j,-j})^{-1}A_{-j,j}=-G_{-j,j}/G_{jj}$. Equivalently, the matrix $M_j\triangleq G-G_{:,j}G_{j,:}/G_{jj}$ has zero row $j$ and zero column $j$, and its remaining block is $(A_{-j,-j})^{-1}$.

 \textbf{The update of $G$.} By \labelcref{eq:principal}, the inverse Gram matrix of $S\setminus\{j\}$ is $(A_{-j,-j})^{-1}$, which by \labelcref{eq:blockinverse} is the second update in \labelcref{eq:obs}.

 \textbf{The update of $B$.} Since $G_{j,:}P_S^\top Y=B_{j,:}$,
 \begin{equation*}
     M_jP_S^\top Y=B-\frac{G_{:,j}B_{j,:}}{G_{jj}} .
 \end{equation*}
 Row $j$ of this matrix is zero because row $j$ of $M_j$ is zero. Because column $j$ of $M_j$ is zero as well, the rows $-j$ of $M_jP_S^\top Y$ equal $(M_j)_{-j,-j}P_{S\setminus\{j\}}^\top Y=(A_{-j,-j})^{-1}P_{S\setminus\{j\}}^\top Y$, which by \labelcref{eq:principal} is the minimizer \labelcref{eq:ridge-solution} on the reduced support. These rows are $B_{-j,:}-G_{-j,j}B_{j,:}/G_{jj}$, the first update in \labelcref{eq:obs}. The updated coefficients are thus the exact minimizer of \labelcref{eq:decoupled} on $S\setminus\{j\}$, not an approximation to it.

 \textbf{The saliency.} Applying \labelcref{eq:Fvalue} on both supports, the term $\tr(Y^\top Y)$ cancels in $\mathcal F(S\setminus\{j\})-\mathcal F(S)$. Since $M_j$ vanishes on row and column $j$, $P_{S\setminus\{j\}}(A_{-j,-j})^{-1}P_{S\setminus\{j\}}^\top=P_SM_jP_S^\top$, and therefore
 \begin{equation*}
  \begin{aligned}
     s_j&=\mathcal F(S\setminus\{j\})-\mathcal F(S)
     =\tr\big(Y^\top P_S(G-M_j)P_S^\top Y\big)
     =\frac{\tr\big(Y^\top P_SG_{:,j}G_{j,:}P_S^\top Y\big)}{G_{jj}}\\
     &=\frac{\lVert G_{j,:}P_S^\top Y\rVert_2^2}{G_{jj}}
     =\frac{\lVert B_{j,:}\rVert_2^2}{G_{jj}},
    \end{aligned}
 \end{equation*}
 which is \labelcref{eq:saliency}.

 \textbf{Interpretation.} A coefficient matrix with $B'_{j,:}=0$ has the same objective value on $S$ as its rows $-j$ have on $S\setminus\{j\}$, since row $j$ multiplies column $j$ of $P_S$ and contributes nothing to the ridge penalty, and every coefficient matrix on $S\setminus\{j\}$ arises in this way. Hence $\mathcal F(S\setminus\{j\})=\min\{\lVert Y-P_SB'\rVert_F^2+\lambda\lVert B'\rVert_F^2:B'_{j,:}=0\}\geq\mathcal F(S)$, and $s_j$ is the smallest increase of the objective over all coefficient matrices with zero row $j$. The objective is a quadratic in $B$ with Hessian $2A$ and zero gradient at the minimizer, so $\lVert Y-P_S(B+\Delta)\rVert_F^2+\lambda\lVert B+\Delta\rVert_F^2=\mathcal F(S)+\tr(\Delta^\top A\Delta)$ for every $\Delta$. Setting row $j$ to zero without changing the other rows, $\Delta=-e_jB_{j,:}$, increases the objective by $A_{jj}\lVert B_{j,:}\rVert_2^2$. Since $A_{jj}G_{jj}\geq1$ for positive definite $A$ (Cauchy--Schwarz in the inner product defined by $A$: $1=(e_j^\top e_j)^2\leq(e_j^\top Ae_j)(e_j^\top A^{-1}e_j)$), the compensated cost $s_j$ is at most the uncompensated one, with equality if and only if $p_j$ is orthogonal to all other retained columns. By \labelcref{eq:blockinverse} and \labelcref{eq:Fvalue} with the target $p_j$,
 \begin{equation*}
     \frac{1}{G_{jj}}
     =\lambda+\lVert p_j\rVert_2^2-p_j^\top P_{S\setminus\{j\}}\big(P_{S\setminus\{j\}}^\top P_{S\setminus\{j\}}+\lambda I\big)^{-1}P_{S\setminus\{j\}}^\top p_j
     =\lambda+\min_{c}\ \lVert p_j-P_{S\setminus\{j\}}c\rVert_2^2+\lambda\lVert c\rVert_2^2 ,
 \end{equation*}
 the ridge-regularized residual of regressing the attention column of entry $j$ on the attention columns of the other retained entries. The saliency is therefore the squared coefficient magnitude of entry $j$ times the part of its attention column that the other retained columns do not reproduce, plus the ridge. An entry with a large coefficient is cheap to delete when its attention column lies close to the span of the other columns, and an entry with a small coefficient is expensive to delete when its attention column lies far from that span.

 \textbf{Deleting a block.} Let $J\subseteq S$ and $R\triangleq S\setminus J$. The block equations of $AG=I$ on the partition $(J,R)$ read $A_{R,J}G_{J,J}+A_{R,R}G_{R,J}=0$ and $A_{R,J}G_{J,R}+A_{R,R}G_{R,R}=I$, and $G_{J,J}$ is positive definite as a principal submatrix of $G$. Eliminating $A_{R,J}=-A_{R,R}G_{R,J}G_{J,J}^{-1}$ from the second equation gives $A_{R,R}\big(G_{R,R}-G_{R,J}G_{J,J}^{-1}G_{J,R}\big)=I$, so $(A_{R,R})^{-1}=G_{R,R}-G_{R,J}G_{J,J}^{-1}G_{J,R}$, and the matrix $M_J\triangleq G-G_{:,J}G_{J,J}^{-1}G_{J,:}$ vanishes on the rows and columns in $J$. The argument for one entry applies verbatim with $G_{J,:}P_S^\top Y=B_{J,:}$: the rows $R$ of $M_JP_S^\top Y$, which are $B_{R,:}-G_{R,J}G_{J,J}^{-1}B_{J,:}$, are the minimizer of \labelcref{eq:decoupled} on $R$, and $(A_{R,R})^{-1}$ is its inverse Gram matrix, which is \labelcref{eq:block}. The exact cost of the joint deletion is
 \begin{equation}
     \mathcal F(S\setminus J)-\mathcal F(S)
     =\tr\big(Y^\top P_S(G-M_J)P_S^\top Y\big)
     =\tr\big(B_{J,:}^\top G_{J,J}^{-1}B_{J,:}\big).
     \label{eq:blockcost}
 \end{equation}
 Since the minimizer on $R$ does not depend on the order in which $R$ is reached, \labelcref{eq:block} gives the same coefficients and inverse as $|J|$ successive single deletions of the elements of $J$. The block cost \labelcref{eq:blockcost} equals $\sum_{j\in J}s_j$ when $G_{J,J}$ is diagonal and differs from it otherwise: it is the sum of the sequential saliencies, each evaluated after the preceding deletions. \textsc{KV-Surgeon} selects $J$ by the individual saliencies, computed before any entry of $J$ is removed. This choice of $J$ is the only approximation of the block step; the updates \labelcref{eq:block} are exact for the surrogate.
 
 \subsection{The single-deletion score}
 \label{sec:app-loo}
 
 The score of \labelcref{eq:loo} is the one case in which the exact loss of \labelcref{eq:family} has a closed form: a single deletion with every other cache parameter at its original value, $S=\{1,\ldots,T\}\setminus\{j\}$, $w=\one$, and $\widehat v_k=v_k$ for $k\in S$. Then $U=V_S$, $\mu_i=1-P_{ij}$, and $\sum_{k\in S}P_{ik}v_k-\mu_io_i=(o_i-P_{ij}v_j)-(1-P_{ij})\,o_i=-P_{ij}(v_j-o_i)$, so \labelcref{eq:compactedattention,eq:family} give
 \begin{equation*}
     \widehat o_i-o_i=-\frac{P_{ij}}{1-P_{ij}}\,(v_j-o_i),
     \qquad
     \mathcal{L}=\sum_{i=1}^{N}\frac{P_{ij}^2}{(1-P_{ij})^2}\lVert v_j-o_i\rVert_2^2=s_j^{\mathrm{LOO}} .
 \end{equation*}
 \citet{goel2026caote} and \citet{an2026restkvrobustkvcache} derive the per-query error for serving-time eviction, with queries read from the prompt suffix; summing its squared norm over the calibration queries carries it to the offline setting. The error is the product of two factors. The renormalization factor $P_{ij}/(1-P_{ij})$ is large for a heavy hitter \citep{zhang2023h2o} with $P_{ij}$ near one, because the mass of the deleted entry is redistributed over retained entries that hold little mass themselves. The value novelty $v_j-o_i$ is small when the entry carries what the head already retrieves, so even a heavily attended entry can be deleted at no cost if its value is close to the current output. \textsc{KV-Surgeon} uses the score in three roles: as the prefilter that forms the candidate pool, as the selection rule of the mild-compaction fallback, and, followed by the same final refit, as the controlled baseline LOO of \cref{sec:exp-obs}. Its second-order truncation, the diagonal $H_{jj}$ of \cref{sec:app-taylor}, drops the factor $(1-P_{ij})^{-2}$.

 \section{Algorithm and implementation}
 \label{sec:app-algimpl}
 \label{sec:app-implementation}
 \Cref{alg:kvsurgeon} in \cref{sec:algdetails} gives the complete procedure for one attention head. This section provides the hyperparameters, solvers, and computational details.

 \textbf{Multipliers as logit biases.} For every retained entry $j\in S$, the compacted cache stores the original key $k_j$, the refit value $\widehat v_j$, and the logarithm $\log w_j$ of its multiplier. At inference, $\log w_j$ is added to the logit of entry $j$ before the softmax, so that the entry contributes $e^{Z_{ij}+\log w_j}=w_je^{Z_{ij}}$ to the unnormalized attention of query $i$, which is \labelcref{eq:compactedattention}. The bound $e^{-\beta}\leq w_j\leq e^{\beta}$ of the final refit is the bound $\lvert\log w_j\rvert\leq\beta$ on the logit bias of any single entry; it bounds the shift that compaction introduces and leaves the original logit $Z_{ij}$ unchanged. Mass preservation, $\mu_i\approx1$, keeps the total unnormalized attention of the compacted cache at that of the original cache, so that the compacted cache draws the same share of attention when it is served next to other tokens (\cref{sec:construction}).

 \textbf{Candidate pool and ridge.} The pool consists of the $m=\min\{T,\lceil ct\rceil\}$ entries with the largest scores \labelcref{eq:loo}, with $c=4$. The score is evaluated with a floor, $P_{ij}^2/((1-P_{ij})^2+10^{-3})$, which bounds the weight of entries that receive almost all attention of a query; the LOO baseline uses the same evaluation. The minimum matters for heads with $T/c<t\leq\tau T$, whose pool is the whole cache. The ridge is set relative to the scale of the Gram matrix on the pool, $\lambda=\lambda_0\tr(P_S^\top P_S)/m$ with $\lambda_0=0.01$, computed once when the pool is formed and kept fixed for all later supports. A fixed $\lambda$ is required for the updates \labelcref{eq:obs,eq:block} to be exact, because the Gram matrix of a reduced support is a principal submatrix of the current one only if both carry the same ridge (\cref{sec:app-obsproof}). Since $m$ can exceed $N$, $P_S^\top P_S$ is singular on the pool in general, and the ridge is what makes $G$ well defined there. The inverse $G$ and the coefficients $B=GP_S^\top Y$ are computed explicitly once, on the pool, from a Cholesky factorization of $P_S^\top P_S+\lambda I$; if the factorization fails, $\lambda$ is increased tenfold, at most three times.

 \textbf{Block deletions.} Each elimination step deletes the $n_b=32$ entries with the smallest saliencies; the last step deletes $\min\{n_b,|S|-t\}$ entries, so that the loop ends at $|S|=t$ exactly. The updates \labelcref{eq:block} require solving linear systems with the $n_b\times n_b$ matrix $G_{J,J}$, which is positive definite as a principal submatrix of $G$. Compared with single deletions, the block step computes the saliencies once per block rather than once per deletion and runs the updates as matrix products with $n_b$ columns; the arithmetic count is the same.

 \textbf{Final refit.} After the elimination, $B$ and $G$ are discarded, and the multipliers and values are solved on $S$ in two ordered steps, following the refit of attention matching \citep{zweiger2026fast}. The multipliers solve the box-constrained least-squares problem
 \begin{equation}
     w\triangleq\argmin_{e^{-\beta}\leq w_j\leq e^{\beta}}\ \lVert\one-P_Sw\rVert_2^2,
     \qquad\beta=3 .
     \label{eq:finalweights}
 \end{equation}
 The box restores the positivity dropped in \cref{sec:surrogate} and bounds the logit bias of every entry. The solver is that of \citet{zweiger2026fast} and solves \labelcref{eq:finalweights} approximately. It works with the unnormalized attention $E=\exp(QK^\top/\sqrt{d_k}-\max)$, the row maximum of the logits subtracted, and its row sums $s$, so that $P=\diag(s)^{-1}E$, and minimizes $\lVert E_Sw-s\rVert_2^2$, which is \labelcref{eq:finalweights} with row $i$ scaled by $s_i$. The unconstrained least-squares solution is computed by a QR solve, clamped to the box, and then refined by two projected gradient steps with step size $1/\lVert E_S\rVert_2^2$, clamping after each step; the spectral norm is estimated by three power iterations. With $w$ fixed, the normalized attention matrix of the compacted cache is $\widehat P\triangleq\diag(P_Sw)^{-1}P_S\diag(w)$, cf.\ \labelcref{eq:compactedattention}, and the values solve the linear least-squares problem
 \begin{equation}
     \widehat V\triangleq\argmin_{\widehat V}\ \lVert\widehat P\widehat V-O\rVert_F^2 ,
     \label{eq:finalvalues}
 \end{equation}
 which is $\min_{\widehat V}\mathcal L(S,w,\widehat V)$, the exact output-preservation loss \labelcref{eq:family} for the chosen support and multipliers. All $d_v$ columns of $\widehat V$ share the Gram matrix $\widehat P^\top\widehat P$ and are solved at once. They are computed by a least-squares solve; if the solve fails numerically, the normal equations $(\widehat P^\top\widehat P+10^{-6}\bar a I)\,\widehat V=\widehat P^\top O$ are solved by a Cholesky factorization instead, where $\bar a$ is the mean diagonal of $\widehat P^\top\widehat P$. The order of the two solves is forced: $\widehat P$ depends on $w$ through $\diag(w)$ and through $P_Sw$, so the values have to be solved after the multipliers. The minimizer of \labelcref{eq:finalvalues} is unique if and only if $P_S$ has full column rank, which cannot hold for $t>N$.

 \textbf{Mild-compaction fallback.} Heads whose budget exceeds the fraction $\tau=0.3$ of their cache, $t>\tau T$, skip the pool, the elimination, and the final refit: they retain the $t$ entries with the largest scores \labelcref{eq:loo}, their original values, and unit multipliers. Eviction is near-lossless at such mild compaction, the $O(t^3)$ solves are largest for exactly these heads, and for $t>N$ the value solve \labelcref{eq:finalvalues} is underdetermined. Disabling the fallback leaves accuracy unchanged and adds $37\%$ to pipeline time (\cref{sec:app-abl-design}).

 \textbf{Complexity.} Forming the Gram matrix of a pool of $m$ candidates costs $O(Nm^2)$ and its inverse $O(m^3)$; the right-hand side $P_S^\top Y$ costs $O(Nmd_v)$ and the initial coefficients $B=GP_S^\top Y$ cost $O(m^2d_v)$. In each elimination step, the saliencies cost $O(|S|(d_v+1))$ given $B$ and the diagonal of $G$, the update of $B$ costs $O(|S|n_b(d_v+1)+n_b^3)$, and the update of $G$ costs $O(|S|^2n_b)$. Summed over the $(m-t)/n_b$ steps, the loop costs $O(m^3+m^2d_v)$, the same count as $m-t$ single deletions. The elimination on a pool of $m$ entries thus costs $O(Nm^2+m^3+Nmd_v+m^2d_v)$ per head, and its memory is dominated by the $m\times m$ matrix $G$. The pool bounds $m$ by $ct$, so these costs grow with the budget $t$ and not with the cache size $T$; eliminating from the full cache instead changes accuracy within noise at $3.8\times$ the compaction time (\cref{sec:app-abl-design}). The fallback removes the $O(t^3)$ terms for the heads with the largest budgets.

 \textbf{Hyperparameters.} All hyperparameters were fixed once on Qwen3-4B with QuALITY at $50\times$ and reused unchanged for every other model, benchmark, and ratio (\cref{sec:expdetails}): the mass weight $\gamma_0=1$ (\cref{sec:gamma}), the relative ridge $\lambda_0=0.01$, the pool factor $c=4$, the block size $n_b=32$, the fallback threshold $\tau=0.3$, and the multiplier bound $\beta=3$. \Cref{sec:app-abl-design} varies $\gamma_0$, $c$, and $\tau$.

 \section{Experimental details}
 \label{sec:expdetails}
 
 This section records what is needed to reproduce and read the experiments; \cref{sec:exp-setup} gives the summary. \Cref{sec:app-protocol} states the protocol and conventions and \cref{sec:app-timing} breaks down where the pipeline time goes.
 
 \subsection{Protocol and conventions}
 \label{sec:app-protocol}
 
 \paragraph{Task and benchmarks.} The model reads a context once, the resulting cache is compacted once by a target factor between $5\times$ and $100\times$, and all benchmark questions about that context are answered from the compacted cache, so calibration and compaction are one-time costs per context. QuALITY uses the dev split, articles 0--49, with $n=894$ multiple-choice questions; one article is one context. LongHealth consists of 20 fictional patient records, grouped into four contexts of five patients with roughly 60k tokens each; each record is compacted as its own chunk at the same ratio and the compacted chunks are concatenated (\cref{sec:construction}), and all $n=400$ multiple-choice questions are evaluated against their contexts. Accuracy is the percentage of correctly answered questions. Answers are extracted from the generated text, and questions whose answer cannot be parsed count as incorrect; parse rates are checked per run and are uniformly high in the reported cells, so parsing does not drive the reported differences. All eight model and benchmark combinations are reported.
 
 \paragraph{Compaction ratio and per-head budgets.} The compaction factor, $5\times$ to $100\times$, is the nominal ratio of the original cache size to the compacted one. Cache-side methods are reported at the nominal ratio because measured effective ratios agree with it to within one percent. Heads differ widely in how much compaction they tolerate, so the compaction factor is a global budget rather than a per-head prescription. Head $(\ell,h)$ receives its own retained fraction $r_{\ell h}$, subject to the constraint that $\operatorname{mean}_{\ell h}r_{\ell h}$ equals the reciprocal of the compaction factor; each head is additionally capped at $r_{\ell h}\leq 0.75$, and the budget a capped head cannot use is redistributed to the other heads in proportion to their shares, so the global ratio is preserved. The allocation is computed once per model on contexts disjoint from the benchmark contexts, following the procedure introduced by \citet{zweiger2026fast}. We measure per head how the log-perplexity of model-generated reference answers degrades as that head alone is compressed, and we allocate the budget along these influence curves. The solver is ratio-agnostic: it finds one set of per-head proportions that works across all target ratios, and this single allocation is rescaled to each compaction factor. Every method in a table consumes the same allocation, so differences between rows are attributable to selection and refit alone; the published configuration of KVzip, reported only in \cref{tab:abl-regime}, is the deliberate exception. Uniform per-head budgets serve as an ablation (\cref{sec:app-ablations}).
 
 \paragraph{Calibration tiers.} All implemented methods score positions against calibration queries recorded from the model itself and, except AM-OMP, come in two tiers that differ mainly in how these queries are generated. The fast tier prompts the model to reproduce the context and records the query vectors of that single pass. The self-study tier additionally generates self-study conversations about the context, which cost more time but match the future workload better. AM-HAK-fast and \textsc{KV-Surgeon}-fast are the fast variants of AM-HAK and \textsc{KV-Surgeon}. AM-OMP, AM-OMP-CS, and \textsc{KV-Surgeon} are self-study methods; AM-OMP-CS is the cheap-solver setting of \citet{zweiger2026fast}, named AM-OMP-fast there, which selects several keys per greedy step and refits less often. After selecting which positions to keep, every calibrated method applies the same support-fixed refit of the retained multipliers and values (\cref{sec:refit}), so accuracy differences between them reflect the selection rule. AM-OMP is evaluated only at the most aggressive ratios because its runtime grows steeply with the retention budget.
 
 \paragraph{Calibration queries.} Self-study generates five conversations per context with vLLM: a verbatim repetition of the context, a summary, an aggregation of key facts, a JSON structuring of the context, and three self-written questions with answers. The repetition is teacher-forced, with the context itself as the answer; the summary, aggregation, and JSON conversations are answered without thinking (at most $2048$ tokens, temperature $0.7$, top-$p$ $0.8$, top-$k$ $20$); the question conversation uses thinking for both writing and answering the questions (at most $8192$ and $4096$ tokens, temperature $0.6$, top-$p$ $0.95$, top-$k$ $20$). The calibration queries are the query vectors of the answer tokens, capped at $50{,}000$ per attention head by random subsampling. The repeat pass is the repetition conversation alone, that is, a single forward pass over the context, the repeat prompt, and a second copy of the context, without any generation. Queries are extracted on-policy, following \citet{zweiger2026fast}: layer $0$ is compacted with the original queries, and for each later layer the same sequences are run through the model with the already compacted earlier layers in place, so every layer is compacted against the queries it receives at inference.
 
 \paragraph{Replication and uncertainty.} Each cell is the mean over five replicate runs with explicit seeds, and the grey number is the standard error over these runs; a single run is noisy to about $\pm1.5$ points at $n=894$ and $\pm2.0$ at $n=400$, comparable to many method gaps, so we only trust orderings that survive the seed error bars. These standard errors measure generation noise under fixed contexts and questions, not uncertainty over the population of contexts, and LongHealth contributes only four composite contexts; generalization rests on the spread of the margins across the eight combinations in \cref{sec:app-grids}. Full-context ceilings come from seed-replicated runs without compaction under the same protocol.
 
 \paragraph{Timing conventions.} Pipeline time is the average wall-clock per context spent on calibration query generation plus compaction, in seconds. It excludes prefilling the context to extract its cache, answer generation, and answer extraction, all of which are common to all methods. Query-free methods therefore report compaction time alone. All runs are single-job SLURM runs on machines with H200 GPUs, launched with the same tooling, so times are comparable within a table; comparisons across models should be treated as indicative. Every table splits calibrated methods by calibration tier, repeat-pass above self-study; AM-OMP-CS uses self-study queries and sits in the self-study group.
 
\paragraph{Defaults.} The method defaults were fixed once on Qwen3-4B with QuALITY at $50\times$ and reused unchanged for every other model, benchmark, and ratio: the mass-term scale $\gamma_0=1$, so that $\gamma$ is the RMS norm of the full-cache outputs; the ridge $\lambda_0=0.01$, applied relative to the mean diagonal of $P_S^\top P_S$ on the candidate pool rather than as an absolute penalty; the block size $n_b=32$ deletions per elimination step; the candidate pool of $m=\min\{T,\lceil 4t\rceil\}$ entries; the threshold $\tau=0.3$ on the retained fraction $t/T$ above which a head is left to plain eviction; and, in the final refit, the bound $|\log w_j|\leq\beta=3$ on the served logit biases, that is, $w_j\in[e^{-3},e^{3}]$, the interval that \cref{tab:abl-gates} refers to. \Cref{sec:app-ablations} varies $\gamma_0$, the pool, and the threshold; \cref{sec:app-implementation} describes how each hyperparameter enters the algorithm.
 
 \subsection{Where the time goes}
 \label{sec:app-timing}
 
 \paragraph{Query generation sets the cost of the self-study tier.} Self-study query generation is a sampled generation pass over the context and costs the same whichever method consumes the queries: $29$ to $33$ seconds per QuALITY article and $193$ to $341$ per LongHealth context on the dense models, $3.4$ to $5.8$ ms per token, and $162$ and $1062$ seconds on Qwen3-30B-A3B, whose mixture-of-experts decoding runs four to five times slower per token under vLLM. A repeat pass, including re-extraction of its query vectors, costs $5$ to $160$ seconds at $50\times$ on the dense models and on Qwen3-30B-A3B with QuALITY, and about $300$ seconds on Qwen3-30B-A3B with LongHealth; the query-free methods need no pass at all. \Cref{tab:timing} splits the self-study pipeline for LOO, \textsc{KV-Surgeon}, and AM-HAK, which consume the same queries. At $50\times$, compaction is $3$ to $25\%$ of the pipeline for all three, so the spread of their pipeline times in \cref{sec:app-grids}, $2$ to $12\%$, comes entirely from the compaction step, while the order-of-magnitude gaps are between tiers.
 
 \paragraph{Compaction grows with the retained fraction.} The compaction step of \textsc{KV-Surgeon} costs $1.4\times$ that of LOO at $50\times$ and $1.8$ to $2.3\times$ at $5\times$, because the block solve grows with the candidate pool: the log-log slope of compaction time against the retained fraction is $0.45$ to $0.6$ for LOO and AM-HAK and $0.65$ to $0.95$ for \textsc{KV-Surgeon}, and since query generation does not depend on the compaction factor, the compaction share rises from about $20\%$ to $50$--$60\%$ across the grid on the dense models. Gemma-3-12B's compaction is small because five of every six layers use sliding-window attention and only the global layers are compacted.
 
 \begin{table}[h]
   \centering
   \footnotesize
   \begin{tabular}{@{}lrrrrrrr@{}}
     \toprule
     & Query gen. & \multicolumn{2}{c}{LOO} & \multicolumn{2}{c}{\textsc{KV-Surgeon}} & \multicolumn{2}{c}{AM-HAK} \\
     \cmidrule(lr){3-4} \cmidrule(lr){5-6} \cmidrule(lr){7-8}
     Model, benchmark & (s) & $50\times$ & $5\times$ & $50\times$ & $5\times$ & $50\times$ & $5\times$ \\
     \midrule
     Qwen3-4B, QuALITY          & 33   & 7.7 & 29  & 10.5 & 53  & 7.0 & 28 \\
     Llama-3.1-8B, LongHealth   & 274  & 66  & 284 & 93   & 578 & 57  & 277 \\
     Gemma-3-12B, QuALITY       & 29   & 1.2 & 4.8 & 1.7  & 11  & 1.1 & 4.8 \\
     Qwen3-30B-A3B, QuALITY     & 162  & 5.5 & 20  & --   & --  & 4.6 & 20 \\
     Qwen3-30B-A3B, LongHealth  & 1062 & 47  & --  & 65   & --  & --  & -- \\
     \bottomrule
   \end{tabular}
   \caption{Where self-study pipeline time goes. Query generation is the sampled self-study pass, in seconds per context; it does not depend on the method or on the compaction factor. The remaining columns are the compaction step alone, in seconds per context, at $50\times$ and $5\times$ for the three methods that consume the same queries. Means over all seeded runs; the sums of query generation and compaction differ from the pipeline times of \cref{sec:app-grids} by up to $15\%$, which is run-to-run timing variation. `--' marks cells not measured.}
   \label{tab:timing}
 \end{table}
 
 \section{Full results}
 \label{sec:app-grids}
 
 The tables that follow cover one model and benchmark combination each and report QA accuracy and pipeline time for every method at all five compaction ratios, under the protocol of \cref{sec:expdetails}. The conventions are the same throughout and are stated once here. Rows are grouped as in \cref{tab:main-acc}, query-free, repeat-pass, and self-study, separated by midrules, and the best accuracy per ratio within each group is bold; an entry is underlined when it trails the bold entry by at most two standard errors of the difference, computed from the two seed standard errors, so that bold without any underlined entry in the same group and column marks a separation beyond seed noise. Time is pipeline time per context in seconds; LongHealth contexts are about eight times longer than QuALITY articles, so every method costs roughly an order of magnitude more there. `--' marks ratios a method does not reach: AM-OMP and AM-OMP-CS are capped because their solver cost grows with the retention budget. Summarization replaces the context by a model-generated summary of it. We use three prompts that lead to summaries of different lengths; the summary length is not a size parameter, so summarization has no table row, and \cref{tab:main-acc,fig:frontier-intro} report the prompt whose resulting cache size was closest to $50\times$ compaction.
 
 \paragraph{Observations across the grids.} Three regularities are not visible in \cref{tab:main-acc}. First, mild compaction saturates: averaged over the eight combinations, the self-study methods span $54.7$ to $59.3$ points at $100\times$ but only $72.9$ to $73.7$ at $5\times$ against a ceiling of $74.0$, and Llama-3.1-8B on QuALITY reaches its ceiling of $62.7\%$ at $20\times$, so that combination is informative only at $50\times$ and $100\times$. Second, \textsc{KV-Surgeon} is never behind beyond noise at the cell level: it is bold in $32$ of the $40$ self-study cells and within noise of the best in the other eight, and its largest deficit anywhere is $1.3$ points, to an AM-OMP variant costing far more. Third, \textsc{KV-Surgeon}-fast is bold or within noise in $37$ of the $40$ repeat-pass cells and separated from every other repeat-pass method in all $16$ cells at $50\times$ and $100\times$; its three losses beyond noise are all at $10\times$ and $5\times$ and at most $1.8$ points.
 
 \paragraph{Query-free baselines.} Q-Filters, StreamingLLM, and the $L_2$ key-norm heuristic use the same optimized per-head budgets as every other method and need no calibration pass. Q-Filters is the best of them at every ratio, with a mean of $46.5\%$ at $50\times$ over the seven combinations that report all three, where \textsc{KV-Surgeon}-fast reaches $63.8\%$ on the same seven for about forty more seconds per context, so at this compaction, calibration is necessary.

 \begin{table}[p]
   \centering
   \footnotesize
   \setlength{\tabcolsep}{3.5pt}
   \begin{tabular}{@{}lcccccccccc@{}}
     \toprule
     & \multicolumn{5}{c}{Accuracy (\%)} & \multicolumn{5}{c}{Pipeline time (s)} \\
     \cmidrule(lr){2-6} \cmidrule(l){7-11}
     Method & $100\times$ & $50\times$ & $20\times$ & $10\times$ & $5\times$ & $100\times$ & $50\times$ & $20\times$ & $10\times$ & $5\times$ \\
     \midrule
     $L_2$ key-norm & 37.0\se{0.3} & 37.9\se{0.4} & 41.2\se{0.6} & 51.1\se{0.3} & 64.2\se{0.3} & 0.7 & 0.6 & 0.5 & 0.6 & 0.6 \\
     Q-Filters & \textbf{47.1}\se{0.7} & \textbf{52.4}\se{0.5} & \textbf{62.0}\se{0.2} & \textbf{66.5}\se{0.3} & \textbf{69.9}\se{0.2} & 0.5 & 0.6 & 0.5 & 0.5 & 0.6 \\
     StreamingLLM & 40.4\se{0.6} & 44.7\se{0.4} & 52.0\se{0.4} & 58.4\se{0.4} & 64.2\se{0.2} & 0.7 & 0.5 & 0.5 & 0.8 & 0.7 \\
     \midrule
     KVzip & 48.0\se{0.4} & 49.8\se{0.4} & 61.6\se{0.4} & 69.1\se{0.5} & 70.5\se{0.2} & 16 & 17 & 21 & 26 & 40 \\
     H2O & 60.0\se{0.5} & 62.3\se{0.4} & \underline{68.2}\se{0.4} & \underline{70.3}\se{0.4} & 71.3\se{0.2} & 16 & 17 & 20 & 27 & 40 \\
     OBCache & 60.1\se{0.3} & 62.4\se{0.7} & \underline{68.3}\se{0.6} & \underline{70.2}\se{0.3} & 71.3\se{0.1} & 16 & 17 & 22 & 27 & 41 \\
     KVMerger & 53.7\se{0.4} & 57.2\se{0.5} & 63.7\se{0.4} & 68.6\se{0.1} & 70.3\se{0.2} & 36 & 42 & 55 & 76 & 106 \\
     AM-HAK-fast & 56.7\se{0.3} & 58.1\se{0.5} & 66.5\se{0.3} & \underline{70.8}\se{0.5} & \textbf{71.9}\se{0.1} & 15 & 17 & 20 & 28 & 38 \\
     \textsc{KV-Surgeon}-fast & \textbf{62.4}\se{0.1} & \textbf{65.9}\se{0.4} & \textbf{68.9}\se{0.3} & \textbf{71.3}\se{0.5} & \textbf{71.9}\se{0.2} & 14 & 16 & 22 & 27 & 28 \\
     \midrule
     AM-HAK & 59.9\se{0.2} & 64.1\se{0.9} & 69.0\se{0.2} & \underline{71.5}\se{0.2} & \underline{71.7}\se{0.4} & 35 & 37 & 41 & 61 & 65 \\
     AM-OMP-CS & \underline{62.2}\se{0.6} & 65.7\se{0.4} & \textbf{70.6}\se{0.2} & \textbf{71.8}\se{0.3} & -- & 51 & 84 & 277 & 919 & -- \\
     AM-OMP & \underline{62.4}\se{0.4} & \underline{66.4}\se{0.2} & \underline{70.1}\se{0.2} & -- & -- & 117 & 314 & 1700 & -- & -- \\
     LOO & 61.5\se{0.5} & 65.6\se{0.2} & \underline{70.2}\se{0.2} & \underline{71.4}\se{0.2} & \textbf{72.3}\se{0.3} & 37 & 38 & 43 & 61 & 66 \\
     \textsc{KV-Surgeon} & \textbf{63.4}\se{0.4} & \textbf{66.8}\se{0.3} & \underline{70.2}\se{0.2} & \underline{71.6}\se{0.5} & \textbf{72.3}\se{0.5} & 37 & 41 & 50 & 68 & 91 \\
     \bottomrule
   \end{tabular}
   \caption{QuALITY with Qwen3-4B. Full-context ceiling $72.1 \pm 0.3\%$.}
   \label{tab:acc-qwen4b-quality}
 \end{table}

 \begin{table}[p]
   \centering
   \footnotesize
   \setlength{\tabcolsep}{3.5pt}
   \begin{tabular}{@{}lcccccccccc@{}}
     \toprule
     & \multicolumn{5}{c}{Accuracy (\%)} & \multicolumn{5}{c}{Pipeline time (s)} \\
     \cmidrule(lr){2-6} \cmidrule(l){7-11}
     Method & $100\times$ & $50\times$ & $20\times$ & $10\times$ & $5\times$ & $100\times$ & $50\times$ & $20\times$ & $10\times$ & $5\times$ \\
     \midrule
     $L_2$ key-norm & \textbf{38.2}\se{0.8} & 37.5\se{0.4} & 36.0\se{0.6} & 51.8\se{0.5} & 64.8\se{0.8} & 7.0 & 4.9 & 4.9 & 6.0 & 5.3 \\
     Q-Filters & \underline{37.6}\se{0.4} & \textbf{43.6}\se{0.4} & \textbf{56.2}\se{0.6} & \textbf{69.1}\se{0.5} & \textbf{76.5}\se{0.4} & 4.8 & 4.8 & 5.1 & 5.4 & 5.2 \\
     StreamingLLM & 34.6\se{0.5} & 39.4\se{0.7} & 42.9\se{0.8} & 55.4\se{0.3} & 64.2\se{0.4} & 5.3 & 5.6 & 5.2 & 4.6 & 5.6 \\
     \midrule
     KVzip & 31.8\se{1.0} & 39.9\se{0.8} & 64.6\se{0.6} & 75.9\se{0.3} & 77.7\se{0.4} & 206 & 222 & 278 & 363 & 530 \\
     H2O & 49.6\se{0.6} & 58.4\se{0.4} & 71.2\se{0.4} & 76.2\se{0.9} & \underline{78.5}\se{0.2} & 203 & 222 & 277 & 370 & 529 \\
     OBCache & 52.8\se{0.6} & 62.8\se{0.9} & 71.2\se{0.4} & 76.2\se{0.4} & \underline{77.9}\se{1.0} & 222 & 233 & 277 & 371 & 555 \\
     KVMerger & 35.2\se{0.4} & 45.5\se{0.4} & 65.6\se{0.8} & 76.2\se{0.6} & \underline{78.1}\se{0.8} & 406 & 461 & 598 & 772 & 1122 \\
     AM-HAK-fast & 43.5\se{0.9} & 53.5\se{1.3} & \textbf{73.0}\se{0.8} & \textbf{78.5}\se{0.4} & \textbf{79.3}\se{0.5} & 204 & 215 & 263 & 357 & 530 \\
     \textsc{KV-Surgeon}-fast & \textbf{57.5}\se{0.8} & \textbf{66.9}\se{0.5} & \underline{72.5}\se{0.6} & 76.7\se{0.6} & \underline{78.8}\se{0.5} & 193 & 220 & 297 & 415 & 628 \\
     \midrule
     AM-HAK & 53.7\se{0.4} & 65.0\se{0.8} & 73.7\se{0.8} & 77.2\se{0.5} & \underline{78.3}\se{0.4} & 331 & 349 & 409 & 512 & 696 \\
     AM-OMP-CS & \underline{57.8}\se{1.0} & \underline{65.9}\se{1.1} & \underline{74.5}\se{0.9} & \textbf{79.1}\se{0.5} & -- & 586 & 1091 & 4518 & 16653 & -- \\
     AM-OMP & \textbf{59.3}\se{0.3} & 65.5\se{0.6} & \textbf{76.0}\se{0.5} & -- & -- & 1678 & 4884 & 29489 & -- & -- \\
     LOO & 53.6\se{0.4} & 64.2\se{1.2} & 74.2\se{0.5} & 77.5\se{0.5} & \underline{78.5}\se{0.5} & 353 & 359 & 426 & 551 & 701 \\
     \textsc{KV-Surgeon} & \underline{59.1}\se{0.5} & \textbf{67.8}\se{0.8} & \underline{74.8}\se{0.7} & \underline{77.8}\se{0.7} & \textbf{78.8}\se{0.4} & 359 & 387 & 523 & 749 & 1227 \\
     \bottomrule
   \end{tabular}
   \caption{LongHealth with Qwen3-4B-Instruct-2507. Full-context ceiling $80.5 \pm 0.8\%$.}
   \label{tab:acc-qwen4b-longhealth}
 \end{table}

 \begin{table}[p]
   \centering
   \footnotesize
   \setlength{\tabcolsep}{3.5pt}
   \begin{tabular}{@{}lcccccccccc@{}}
     \toprule
     & \multicolumn{5}{c}{Accuracy (\%)} & \multicolumn{5}{c}{Pipeline time (s)} \\
     \cmidrule(lr){2-6} \cmidrule(l){7-11}
     Method & $100\times$ & $50\times$ & $20\times$ & $10\times$ & $5\times$ & $100\times$ & $50\times$ & $20\times$ & $10\times$ & $5\times$ \\
     \midrule
     $L_2$ key-norm & 38.0\se{0.4} & 40.2\se{0.5} & 48.1\se{0.5} & 56.4\se{0.5} & 60.6\se{0.1} & 0.5 & 0.5 & 0.5 & 0.5 & 0.6 \\
     Q-Filters & \textbf{40.8}\se{0.3} & \textbf{47.7}\se{0.8} & \textbf{56.7}\se{1.0} & \textbf{60.9}\se{0.4} & \textbf{62.5}\se{0.3} & 0.6 & 0.6 & 0.6 & 0.6 & 0.6 \\
     StreamingLLM & 38.7\se{0.5} & 43.2\se{0.4} & 49.0\se{0.7} & 54.1\se{0.4} & 59.2\se{0.2} & 0.5 & 0.5 & 0.6 & 0.6 & 0.6 \\
     \midrule
     KVzip & 46.0\se{0.6} & 50.5\se{0.6} & 58.9\se{0.1} & 64.1\se{0.6} & \underline{63.4}\se{0.5} & 14 & 15 & 18 & 24 & 34 \\
     H2O & 55.7\se{0.5} & 59.1\se{0.4} & \underline{63.1}\se{0.3} & \textbf{65.5}\se{0.3} & \underline{64.2}\se{0.4} & 14 & 15 & 18 & 25 & 34 \\
     OBCache & 56.9\se{0.4} & 59.8\se{0.3} & \underline{63.2}\se{0.4} & 63.7\se{0.1} & \underline{64.1}\se{0.7} & 15 & 16 & 19 & 25 & 35 \\
     KVMerger & 49.4\se{0.6} & 54.0\se{0.7} & 59.6\se{0.4} & 62.4\se{0.3} & \underline{63.4}\se{0.5} & 32 & 37 & 48 & 62 & 92 \\
     AM-HAK-fast & 54.9\se{0.7} & 58.7\se{0.6} & \underline{63.0}\se{0.4} & 63.8\se{0.3} & \textbf{64.5}\se{0.4} & 14 & 15 & 18 & 24 & 34 \\
     \textsc{KV-Surgeon}-fast & \textbf{58.5}\se{0.6} & \textbf{61.0}\se{0.4} & \textbf{63.4}\se{0.6} & 63.8\se{0.6} & \underline{64.3}\se{0.8} & 13 & 15 & 20 & 22 & 29 \\
     \midrule
     AM-HAK & 57.0\se{0.7} & 60.4\se{0.5} & \textbf{63.2}\se{0.4} & \underline{63.9}\se{0.2} & \textbf{63.8}\se{0.2} & 28 & 29 & 33 & 40 & 53 \\
     AM-OMP-CS & 57.1\se{0.6} & \underline{60.6}\se{0.8} & \underline{62.8}\se{0.9} & \textbf{64.6}\se{0.4} & -- & 42 & 74 & 283 & 934 & -- \\
     AM-OMP & \underline{58.6}\se{0.5} & \underline{61.3}\se{0.6} & \underline{63.1}\se{0.7} & -- & -- & 101 & 288 & 1573 & -- & -- \\
     LOO & 58.0\se{0.4} & \underline{61.2}\se{0.7} & \underline{63.1}\se{0.3} & 63.1\se{0.4} & \underline{63.2}\se{0.3} & 27 & 28 & 32 & 39 & 51 \\
     \textsc{KV-Surgeon} & \textbf{60.3}\se{0.7} & \textbf{62.3}\se{0.5} & \textbf{63.2}\se{0.6} & \underline{63.9}\se{0.3} & \underline{63.5}\se{0.5} & 28 & 31 & 39 & 50 & 68 \\
     \bottomrule
   \end{tabular}
   \caption{QuALITY with Llama-3.1-8B-Instruct. Full-context ceiling $62.7 \pm 0.5\%$.}
   \label{tab:acc-llama-quality}
 \end{table}

 \begin{table}[p]
   \centering
   \footnotesize
   \setlength{\tabcolsep}{3.5pt}
   \begin{tabular}{@{}lcccccccccc@{}}
     \toprule
     & \multicolumn{5}{c}{Accuracy (\%)} & \multicolumn{5}{c}{Pipeline time (s)} \\
     \cmidrule(lr){2-6} \cmidrule(l){7-11}
     Method & $100\times$ & $50\times$ & $20\times$ & $10\times$ & $5\times$ & $100\times$ & $50\times$ & $20\times$ & $10\times$ & $5\times$ \\
     \midrule
     $L_2$ key-norm & 33.0\se{1.2} & \underline{34.1}\se{1.6} & 37.6\se{0.7} & \textbf{50.5}\se{0.3} & \textbf{62.5}\se{0.6} & 5.6 & 5.9 & 5.6 & 5.6 & 6.2 \\
     Q-Filters & \underline{36.0}\se{0.8} & \underline{36.1}\se{1.1} & \underline{39.9}\se{0.6} & \underline{49.5}\se{0.8} & \underline{61.4}\se{0.2} & 7.0 & 5.7 & 6.0 & 6.6 & 5.7 \\
     StreamingLLM & \textbf{37.1}\se{1.1} & \textbf{37.2}\se{0.8} & \textbf{42.2}\se{1.1} & 45.4\se{0.5} & 56.3\se{0.8} & 5.9 & 5.2 & 5.7 & 5.6 & 5.7 \\
     \midrule
     KVzip & 23.3\se{1.1} & 41.0\se{1.4} & 63.5\se{0.7} & \underline{71.3}\se{0.4} & 72.7\se{0.5} & 173 & 179 & 229 & 302 & 437 \\
     H2O & 43.5\se{0.7} & 50.6\se{0.9} & 66.2\se{0.6} & 70.2\se{0.5} & \underline{73.2}\se{0.6} & 173 & 189 & 220 & 307 & 437 \\
     OBCache & 44.7\se{0.9} & 52.4\se{0.2} & 66.6\se{0.4} & \underline{71.5}\se{1.0} & \underline{73.2}\se{0.5} & 193 & 202 & 232 & 326 & 457 \\
     KVMerger & 34.5\se{0.8} & 42.8\se{0.6} & 57.3\se{0.2} & 67.8\se{0.6} & 72.5\se{0.5} & 337 & 397 & 505 & 643 & 929 \\
     AM-HAK-fast & 44.4\se{1.2} & 53.5\se{1.1} & \underline{68.0}\se{0.5} & \textbf{73.5}\se{1.4} & \textbf{74.3}\se{0.3} & 165 & 180 & 221 & 303 & 447 \\
     \textsc{KV-Surgeon}-fast & \textbf{53.9}\se{1.0} & \textbf{61.5}\se{0.6} & \textbf{68.8}\se{0.4} & \underline{71.4}\se{0.5} & 73.0\se{0.5} & 153 & 181 & 262 & 287 & 403 \\
     \midrule
     AM-HAK & 48.4\se{1.1} & 58.2\se{0.5} & \underline{70.2}\se{0.7} & \underline{72.8}\se{0.3} & 70.5\se{1.0} & 291 & 311 & 350 & 456 & 614 \\
     AM-OMP-CS & 47.9\se{0.5} & 55.8\se{0.9} & 66.1\se{0.5} & -- & -- & 564 & 1209 & 5867 & -- & -- \\
     AM-OMP & 49.2\se{0.6} & 57.3\se{0.9} & \underline{69.0}\se{0.6} & -- & -- & 1650 & 5292 & 35590 & -- & -- \\
     LOO & 44.5\se{1.7} & 59.1\se{0.6} & 66.8\se{0.7} & 70.8\se{0.9} & \underline{73.4}\se{0.6} & 295 & 323 & 360 & 453 & 620 \\
     \textsc{KV-Surgeon} & \textbf{54.0}\se{0.8} & \textbf{62.1}\se{0.3} & \textbf{71.5}\se{1.2} & \textbf{73.2}\se{0.5} & \textbf{74.1}\se{0.8} & 313 & 349 & 487 & 654 & 908 \\
     \bottomrule
   \end{tabular}
   \caption{LongHealth with Llama-3.1-8B-Instruct. Full-context ceiling $73.0 \pm 0.8\%$.}
   \label{tab:acc-llama-longhealth}
 \end{table}

 \begin{table}[p]
   \centering
   \footnotesize
   \setlength{\tabcolsep}{3.5pt}
   \begin{tabular}{@{}lcccccccccc@{}}
     \toprule
     & \multicolumn{5}{c}{Accuracy (\%)} & \multicolumn{5}{c}{Pipeline time (s)} \\
     \cmidrule(lr){2-6} \cmidrule(l){7-11}
     Method & $100\times$ & $50\times$ & $20\times$ & $10\times$ & $5\times$ & $100\times$ & $50\times$ & $20\times$ & $10\times$ & $5\times$ \\
     \midrule
     $L_2$ key-norm & 41.3\se{0.4} & 42.5\se{0.6} & 45.9\se{0.4} & 52.4\se{0.2} & 68.3\se{0.3} & 0.3 & 0.3 & 0.3 & 0.3 & 0.4 \\
     Q-Filters & \textbf{46.3}\se{0.6} & \textbf{50.9}\se{0.3} & \textbf{59.9}\se{0.7} & \textbf{70.6}\se{0.4} & \textbf{77.3}\se{0.1} & 0.3 & 0.4 & 0.4 & 0.4 & 0.3 \\
     StreamingLLM & 44.7\se{0.5} & 47.5\se{0.3} & 53.8\se{0.1} & 60.9\se{0.4} & 70.2\se{0.3} & 0.3 & 0.3 & 0.3 & 0.3 & 0.3 \\
     \midrule
     KVzip & 48.3\se{0.2} & 53.0\se{0.4} & 60.4\se{0.3} & 70.3\se{0.7} & 78.7\se{0.2} & 56 & 63 & 66 & 96 & 78 \\
     H2O & 60.6\se{0.4} & 66.5\se{0.3} & 73.7\se{0.7} & \underline{78.3}\se{0.2} & \textbf{80.5}\se{0.4} & 53 & 53 & 59 & 64 & 76 \\
     OBCache & 61.7\se{0.2} & 67.3\se{0.4} & 73.1\se{0.5} & \textbf{78.4}\se{0.3} & \underline{80.0}\se{0.3} & 62 & 63 & 61 & 67 & 88 \\
     KVMerger & 58.2\se{0.6} & 63.9\se{0.6} & 72.2\se{0.2} & 75.1\se{0.3} & 79.2\se{0.1} & 76 & 82 & 94 & 100 & 122 \\
     AM-HAK-fast & 60.3\se{0.3} & 65.8\se{0.5} & 71.2\se{0.3} & \underline{77.3}\se{0.4} & \underline{79.5}\se{0.3} & 51 & 52 & 56 & 63 & 77 \\
     \textsc{KV-Surgeon}-fast & \textbf{66.5}\se{0.2} & \textbf{71.4}\se{0.3} & \textbf{76.5}\se{0.2} & \textbf{78.4}\se{0.4} & \underline{80.2}\se{0.3} & 53 & 54 & 61 & 65 & 76 \\
     \midrule
     AM-HAK & 63.5\se{0.5} & 69.5\se{0.3} & 74.0\se{0.3} & 78.2\se{0.3} & \underline{80.3}\se{0.4} & 156 & 162 & 168 & 172 & 194 \\
     AM-OMP-CS & 63.5\se{0.6} & 69.7\se{0.5} & 75.2\se{0.4} & \underline{78.5}\se{0.4} & -- & 165 & 183 & 283 & 616 & -- \\
     AM-OMP & 63.6\se{0.6} & 69.2\se{0.4} & 75.5\se{0.2} & -- & -- & 199 & 300 & 996 & -- & -- \\
     LOO & 64.7\se{0.3} & 69.6\se{0.3} & 74.9\se{0.3} & \underline{78.3}\se{0.7} & 79.8\se{0.1} & 163 & 165 & 170 & 173 & 214 \\
     \textsc{KV-Surgeon} & \textbf{67.6}\se{0.4} & \textbf{72.5}\se{0.6} & \textbf{77.4}\se{0.3} & \textbf{79.2}\se{0.2} & \textbf{80.5}\se{0.3} & 164 & 159 & 166 & 181 & 198 \\
     \bottomrule
   \end{tabular}
   \caption{QuALITY with Qwen3-30B-A3B. Full-context ceiling $80.7 \pm 0.4\%$.}
   \label{tab:acc-qwen30b-quality}
 \end{table}

 \begin{table}[p]
   \centering
   \footnotesize
   \setlength{\tabcolsep}{3.5pt}
   \begin{tabular}{@{}lcccccccccc@{}}
     \toprule
     & \multicolumn{5}{c}{Accuracy (\%)} & \multicolumn{5}{c}{Pipeline time (s)} \\
     \cmidrule(lr){2-6} \cmidrule(l){7-11}
     Method & $100\times$ & $50\times$ & $20\times$ & $10\times$ & $5\times$ & $100\times$ & $50\times$ & $20\times$ & $10\times$ & $5\times$ \\
     \midrule
     Q-Filters & -- & 43.2\se{0.5} & -- & -- & -- & -- & -- & -- & -- & -- \\
     \midrule
     OBCache & 52.7\se{0.4} & 62.8\se{0.4} & 68.6\se{0.5} & \underline{74.4}\se{0.1} & \underline{76.7}\se{0.3} & 391 & 395 & 402 & 409 & 424 \\
     AM-HAK-fast & 50.4\se{1.1} & 59.8\se{0.6} & 66.9\se{0.4} & \underline{74.3}\se{0.5} & \textbf{77.2}\se{0.2} & 358 & 361 & 362 & 372 & 389 \\
     \textsc{KV-Surgeon}-fast & \textbf{54.7}\se{0.3} & \textbf{64.1}\se{0.2} & \textbf{70.2}\se{0.5} & \textbf{75.0}\se{0.3} & \underline{77.1}\se{0.2} & 363 & 366 & 373 & 386 & 413 \\
     \midrule
     AM-HAK & 51.4\se{0.4} & 61.4\se{0.4} & 69.8\se{0.5} & 73.7\se{0.1} & \underline{76.8}\se{0.3} & 1121 & 1129 & 1130 & 1159 & 1162 \\
     AM-OMP-CS & 52.6\se{0.5} & 63.1\se{0.2} & 70.0\se{0.3} & -- & -- & 1271 & 1569 & 3521 & -- & -- \\
     AM-OMP & 53.3\se{0.2} & \underline{63.9}\se{0.6} & -- & -- & -- & 2147 & 4011 & -- & -- & -- \\
     LOO & 52.1\se{0.3} & 62.3\se{0.7} & 69.3\se{0.4} & \underline{74.2}\se{1.2} & \underline{77.0}\se{0.3} & 1109 & 1112 & 1131 & 1143 & 1159 \\
     \textsc{KV-Surgeon} & \textbf{55.9}\se{0.3} & \textbf{65.2}\se{0.4} & \textbf{71.4}\se{0.4} & \textbf{75.8}\se{0.6} & \textbf{77.6}\se{0.3} & 1124 & 1135 & 1149 & 1177 & 1211 \\
     \bottomrule
   \end{tabular}
   \caption{LongHealth with Qwen3-30B-A3B. Full-context ceiling $78.3 \pm 0.5\%$. Of the query-free methods, only Q-Filters at $50\times$ was run on this combination; KVzip, H2O, and KVMerger were not run.}
   \label{tab:acc-qwen30b-longhealth}
 \end{table}

 \begin{table}[p]
   \centering
   \footnotesize
   \setlength{\tabcolsep}{3.5pt}
   \begin{tabular}{@{}lcccccccccc@{}}
     \toprule
     & \multicolumn{5}{c}{Accuracy (\%)} & \multicolumn{5}{c}{Pipeline time (s)} \\
     \cmidrule(lr){2-6} \cmidrule(l){7-11}
     Method & $100\times$ & $50\times$ & $20\times$ & $10\times$ & $5\times$ & $100\times$ & $50\times$ & $20\times$ & $10\times$ & $5\times$ \\
     \midrule
     $L_2$ key-norm & 47.4\se{0.5} & 48.5\se{0.3} & 49.3\se{0.2} & 50.8\se{0.1} & 55.3\se{0.5} & 0.3 & 0.3 & 0.3 & 0.3 & 0.3 \\
     Q-Filters & \textbf{53.2}\se{0.5} & \textbf{55.9}\se{0.3} & \textbf{61.7}\se{0.4} & \textbf{65.4}\se{0.4} & \textbf{68.2}\se{0.4} & 0.3 & 0.3 & 0.3 & 0.3 & 0.3 \\
     StreamingLLM & 49.3\se{0.6} & 50.2\se{0.3} & 52.4\se{0.5} & 55.7\se{0.3} & 61.4\se{0.4} & 0.3 & 0.3 & 0.3 & 0.3 & 0.3 \\
     \midrule
     KVzip & 56.0\se{0.2} & 57.6\se{0.8} & 63.1\se{0.4} & 67.2\se{0.5} & 69.1\se{0.4} & 6.0 & 6.2 & 6.7 & 7.4 & 9.1 \\
     H2O & 61.5\se{0.3} & 66.2\se{0.2} & 68.1\se{0.2} & \textbf{71.1}\se{0.4} & \textbf{71.3}\se{0.4} & 6.0 & 6.1 & 6.6 & 7.4 & 9.0 \\
     OBCache & 62.7\se{0.3} & 66.6\se{0.4} & 68.0\se{0.1} & \underline{70.5}\se{0.4} & \underline{71.2}\se{0.2} & 6.19 & 6.47 & 6.84 & 7.54 & 9.39 \\
     KVMerger & 56.5\se{0.2} & 61.1\se{0.2} & 64.8\se{0.4} & 67.6\se{0.3} & 69.3\se{0.2} & 12 & 12 & 15 & 19 & 25 \\
     AM-HAK-fast & 59.5\se{0.2} & 62.2\se{0.4} & 66.8\se{0.5} & 69.3\se{0.4} & 69.5\se{0.3} & 6.1 & 6.2 & 6.7 & 7.4 & 8.9 \\
     \textsc{KV-Surgeon}-fast & \textbf{65.2}\se{0.3} & \textbf{68.1}\se{0.5} & \textbf{70.4}\se{0.8} & \underline{71.0}\se{0.6} & \underline{70.6}\se{0.2} & 6.0 & 6.5 & 7.4 & 8.8 & 11 \\
     \midrule
     AM-HAK & 61.5\se{0.6} & 65.1\se{0.3} & 68.8\se{0.3} & \underline{70.0}\se{0.3} & \underline{70.9}\se{0.4} & 29 & 30 & 30 & 31 & 34 \\
     AM-OMP-CS & 64.2\se{0.3} & \underline{67.2}\se{0.2} & \underline{69.2}\se{0.3} & \underline{70.3}\se{0.3} & -- & 32 & 37 & 56 & 125 & -- \\
     AM-OMP & \underline{64.7}\se{0.4} & \textbf{67.7}\se{0.3} & \underline{69.8}\se{0.4} & -- & -- & 38 & 59 & 180 & -- & -- \\
     LOO & 60.5\se{0.3} & 65.1\se{0.5} & 68.6\se{0.3} & \underline{70.5}\se{0.4} & \underline{70.4}\se{0.4} & 29 & 29 & 30 & 31 & 33 \\
     \textsc{KV-Surgeon} & \textbf{65.5}\se{0.2} & \underline{67.5}\se{0.7} & \textbf{70.0}\se{0.4} & \textbf{70.9}\se{0.4} & \textbf{71.1}\se{0.2} & 30 & 31 & 33 & 35 & 39 \\
     \bottomrule
   \end{tabular}
   \caption{QuALITY with Gemma-3-12B-it. Full-context ceiling $70.6 \pm 0.4\%$.}
   \label{tab:acc-gemma-quality}
 \end{table}

 \begin{table}[p]
   \centering
   \footnotesize
   \setlength{\tabcolsep}{3.5pt}
   \begin{tabular}{@{}lcccccccccc@{}}
     \toprule
     & \multicolumn{5}{c}{Accuracy (\%)} & \multicolumn{5}{c}{Pipeline time (s)} \\
     \cmidrule(lr){2-6} \cmidrule(l){7-11}
     Method & $100\times$ & $50\times$ & $20\times$ & $10\times$ & $5\times$ & $100\times$ & $50\times$ & $20\times$ & $10\times$ & $5\times$ \\
     \midrule
     $L_2$ key-norm & 37.6\se{0.5} & \underline{38.9}\se{1.2} & \underline{37.8}\se{0.6} & 40.4\se{0.4} & 41.0\se{0.6} & 6.2 & 6.2 & 6.3 & 6.1 & 6.0 \\
     Q-Filters & 38.4\se{0.2} & \underline{39.0}\se{0.7} & \textbf{39.4}\se{0.9} & \textbf{44.5}\se{0.8} & \textbf{53.9}\se{0.7} & 6.0 & 6.1 & 6.3 & 6.1 & 6.1 \\
     StreamingLLM & \textbf{40.9}\se{0.3} & \textbf{39.5}\se{0.9} & \underline{39.3}\se{0.6} & 42.0\se{0.3} & 44.4\se{0.6} & 6.1 & 5.9 & 6.0 & 6.1 & 6.3 \\
     \midrule
     KVzip & 38.1\se{0.5} & 39.8\se{0.7} & 44.5\se{0.6} & 52.9\se{0.8} & 60.7\se{0.6} & 55 & 59 & 64 & 74 & 95 \\
     H2O & 38.5\se{0.6} & 47.3\se{0.5} & 55.1\se{0.2} & 63.5\se{0.6} & \underline{68.8}\se{0.7} & 54 & 60 & 64 & 74 & 91 \\
     OBCache & 41.3\se{0.5} & 50.2\se{0.8} & 56.1\se{0.4} & 64.0\se{0.8} & \underline{69.7}\se{0.5} & 59.2 & 59.9 & 66.4 & 75.1 & 95.3 \\
     KVMerger & 33.7\se{0.6} & 36.1\se{0.8} & 42.5\se{0.6} & 50.1\se{0.9} & 59.2\se{0.6} & 97 & 106 & 140 & 170 & 239 \\
     AM-HAK-fast & 41.5\se{1.0} & 45.6\se{1.2} & 54.6\se{0.8} & 61.8\se{0.6} & 66.0\se{0.7} & 57 & 59 & 80 & 73 & 96 \\
     \textsc{KV-Surgeon}-fast & \textbf{45.6}\se{0.6} & \textbf{52.1}\se{0.4} & \textbf{61.1}\se{0.3} & \textbf{66.5}\se{0.7} & \textbf{71.2}\se{1.1} & 53 & 59 & 71 & 97 & 138 \\
     \midrule
     AM-HAK & 44.7\se{0.7} & 52.7\se{1.4} & \underline{62.5}\se{0.6} & \underline{67.2}\se{0.2} & \underline{71.0}\se{0.5} & 198 & 207 & 210 & 227 & 291 \\
     AM-OMP-CS & 43.6\se{0.9} & 50.9\se{0.8} & \underline{61.2}\se{0.9} & 66.2\se{0.7} & -- & 225 & 287 & 624 & 2048 & -- \\
     AM-OMP & \underline{45.5}\se{1.2} & \underline{54.5}\se{0.9} & \underline{61.9}\se{1.0} & -- & -- & 353 & 689 & 2831 & -- & -- \\
     LOO & 42.5\se{0.7} & 49.6\se{0.7} & 60.6\se{0.6} & \underline{67.2}\se{0.7} & 69.8\se{0.7} & 206 & 205 & 213 & 218 & 249 \\
     \textsc{KV-Surgeon} & \textbf{48.4}\se{1.1} & \textbf{56.4}\se{0.7} & \textbf{63.0}\se{0.7} & \textbf{68.8}\se{0.8} & \textbf{71.8}\se{0.5} & 194 & 201 & 224 & 263 & 346 \\
     \bottomrule
   \end{tabular}
   \caption{LongHealth with Gemma-3-12B-it. Full-context ceiling $74.5 \pm 0.4\%$.}
   \label{tab:acc-gemma-longhealth}
 \end{table}
 
 \subsection{Long Code Arena}
 \label{sec:app-lca}
 
 \Cref{tab:lca-qwen4b} reports project-level code completion on Long Code Arena \citep{bogomolov2024long} with Qwen3-4B-Instruct-2507, the experiment behind the last paragraph of \cref{sec:exp-main}. The benchmark is the \texttt{large\_context} configuration of its project-level code completion task: each datapoint is a repository snapshot at a fixed commit, one held-out completion file, and a list of target lines in that file.
 
 \paragraph{Contexts and questions.} We walk the test split in order and keep the first snapshot of each not-yet-seen repository whose Python files plus completion file fit $120{,}000$ Gemma-3 tokens, until $20$ repositories are collected; the dataset order and commit hashes are recorded, so the selection is deterministic. The context of a repository is the concatenation of its non-empty \texttt{.py} files in snapshot order, each preceded by a header with its path; the completion file itself is not part of the context. The median context is $73{,}298$ Qwen tokens (range about $45$k to $95$k). Every listed target line with non-blank content is one question, $n=650$ in total. A question shows the path of the completion file and its prefix, all lines before the target, in a code fence, and asks for exactly the next line; prefixes longer than $16{,}000$ characters are truncated to their tail at a line boundary, which affects two questions. The prefix is part of the question, not of the compacted context, so compaction is tested through the repository context alone. The answer is generated inside an open code fence with at most $64$ new tokens.
 
 \paragraph{Compaction and scoring.} As for LongHealth (\cref{sec:construction}), each repository is compacted in chunks: consecutive files are packed into chunks of at most $10{,}000$ tokens without crossing file boundaries, about nine per repository, each chunk is compacted at the target ratio, and the compacted chunks are concatenated. Head budgets, calibration, final refit, method configurations, and the five-seed protocol are identical to the LongHealth cells; seeds affect calibration-query sampling and generation, not the fixed set of repositories and questions. A completion is correct when its first generated code line, after dropping empty lines and fence lines and stripping surrounding whitespace, matches the reference line exactly; accuracy is the mean over the $650$ questions. Time is query generation plus compaction per repository. The full-context ceiling under the same questions and scorer is $40.6\%$, and the query-free baselines reach $22$ to $23\%$ from the file prefix alone, so the effective range of the benchmark is about $22$ to $41$ points.
 
 \paragraph{Results.} At $100\times$ no calibrated method is more than $0.4$ points above the best query-free baseline, so the budget rather than the selection rule limits accuracy there. From $50\times$ on, the calibrated methods separate. \textsc{KV-Surgeon} and \textsc{KV-Surgeon}-fast match AM-OMP within uncertainty at $50\times$ with $9.7\times$ and $21.5\times$ less pipeline time, \textsc{KV-Surgeon} has the highest point estimate at $20\times$ and $10\times$, and at $5\times$ every calibrated method except KVzip and KVMerger is within $1.8$ points of the ceiling. OBCache is again the strongest ported method, and AM-OMP-CS is the only prior method within uncertainty of \textsc{KV-Surgeon} at $20\times$, at $6.7\times$ its cost.
 
 \begin{table}[t]
   \centering
   \caption{Code completion accuracy (\%) and pipeline time vs.\ compaction ratio on Long Code Arena project-level code completion with Qwen3-4B-Instruct-2507 (20 Python repository snapshots, $n=650$ next-line completions, exact match of the first generated line). Each entry is the mean over five seeds; the grey number is the standard error over seeds. Time is average query-generation plus compaction wall-clock per repository, in seconds. Full-context ceiling: $40.6\%$. The best accuracy per ratio is bold; underlined entries are within $2\sqrt{\mathrm{se}^2+\mathrm{se}_\mathrm{best}^2}$ of the best. `--' marks ratios the AM-OMP solvers do not reach.}
   \label{tab:lca-qwen4b}
   \small
   \setlength{\tabcolsep}{3.0pt}
   \begin{tabular}{l ccccc ccccc}
   \toprule
   & \multicolumn{5}{c}{Accuracy (\%)} & \multicolumn{5}{c}{Time (s)} \\
   \cmidrule(lr){2-6} \cmidrule(lr){7-11}
   Method & $100\times$ & $50\times$ & $20\times$ & $10\times$ & $5\times$
          & $100\times$ & $50\times$ & $20\times$ & $10\times$ & $5\times$ \\
   \midrule
   StreamingLLM        & 22.6\se{0.2} & 23.3\se{0.2} & 22.0\se{0.1} & 24.5\se{0.1} & 32.5\se{0.1} & 7.62 & 8.07 & 7.78 & 8.18 & 8.39 \\
   $L_2$ key-norm      & \underline{23.2}\se{0.1} & 23.4\se{0.1} & 23.4\se{0.2} & 23.4\se{0.1} & 27.6\se{0.2} & 7.45 & 8.55 & 8.77 & 7.70 & 8.12 \\
   Q-Filters           & 22.8\se{0.1} & 22.4\se{0.2} & 23.8\se{0.1} & 27.4\se{0.1} & 34.7\se{0.2} & 7.11 & 7.41 & 7.50 & 8.02 & 7.82 \\
   KVzip               & 20.5\se{0.2} & 22.4\se{0.2} & 28.6\se{0.4} & 36.2\se{0.5} & 38.3\se{0.2} & 217  & 232 & 292 & 379 & 596 \\
   H2O                 & 22.5\se{0.1} & 24.1\se{0.3} & 30.9\se{0.2} & 36.4\se{0.2} & \underline{39.4}\se{0.2} & 221  & 230 & 307 & 385 & 572 \\
   KVMerger            & 13.5\se{0.2} & 15.1\se{0.2} & 24.4\se{0.1} & 31.7\se{0.2} & 35.9\se{0.2} & 490  & 554 & 688 & 962 & 1293 \\
   OBCache             & 22.0\se{0.2} & 25.4\se{0.3} & 32.7\se{0.2} & 37.2\se{0.2} & \underline{39.2}\se{0.3} & 235  & 243 & 309 & 391 & 554 \\
   AM-HAK-fast         & 22.2\se{0.4} & 23.7\se{0.3} & 32.4\se{0.1} & 37.4\se{0.1} & 39.1\se{0.3} & 222  & 245 & 287 & 384 & 536 \\
   AM-HAK              & 22.7\se{0.3} & 24.6\se{0.5} & 31.0\se{0.2} & 37.2\se{0.3} & 39.1\se{0.2} & 437  & 448 & 505 & 628 & 846 \\
   AM-OMP-CS         & \underline{23.1}\se{0.4} & 25.5\se{0.3} & \underline{33.3}\se{0.3} & --           & --           & 649  & 1110 & 5128 & -- & -- \\
   AM-OMP              & \underline{23.5}\se{0.3} & \textbf{27.0}\se{0.4} & --           & --           & --           & 1646 & 5037 & -- & -- & -- \\
   LOO                 & 22.6\se{0.3} & 25.3\se{0.2} & 32.5\se{0.2} & 37.2\se{0.2} & 38.8\se{0.4} & 444  & 469 & 534 & 661 & 872 \\
   \textsc{KV-Surgeon}-fast     & 22.8\se{0.2} & \underline{26.5}\se{0.2} & 33.1\se{0.3} & 36.9\se{0.2} & \textbf{39.8}\se{0.1} & 204  & 234 & 297 & 407 & 542 \\
   \textsc{KV-Surgeon}          & \textbf{23.6}\se{0.3} & \underline{26.8}\se{0.4} & \textbf{33.9}\se{0.2} & \textbf{38.1}\se{0.3} & \underline{39.7}\se{0.2} & 474  & 519 & 771 & 875 & 1322 \\
   \bottomrule
   \end{tabular}
 \end{table}
 
 \clearpage
 
 \section{Component ablations}
 \label{sec:app-ablations}
 
 This section holds every ablation of the shared pipeline and of \textsc{KV-Surgeon} behind \cref{sec:exp-obs} and \cref{tab:comp-ablation}: the shared pipeline components in \cref{sec:app-abl-shared}, compensation in \cref{sec:app-abl-comp}, and the efficiency and design choices in \cref{sec:app-abl-design}. Every ablation runs on Qwen3-4B with QuALITY at $50\times$, changes one component of the pipeline of \cref{sec:prelim} at a time, and reports the mean over five seeds with one standard error over seeds. Two protocols are used and must not be mixed. The $50$-article protocol (\cref{tab:ablation-main,tab:abl-refit-budget}) is that of the main tables, $894$ questions; its self-study controls are the \textsc{KV-Surgeon}, LOO, and AM-HAK entries of \cref{tab:main-acc}, and its time is pipeline time, calibration-query generation plus compaction. The $25$-article protocol (\cref{tab:ablation-pool}) uses the first $25$ articles, so its accuracies are compared only against the reference row of the same panel and never against the main-table numbers, and its time is compaction only, because query generation is identical across arms. The fast-tier row of \cref{tab:ablation-main} was run with the fast configuration in use at the time of the ablation, whose reference values ($65.5\%$, $5.4$ s) differ from the \textsc{KV-Surgeon}-fast entry of \cref{tab:main-acc}; it too is read only against its own control. Defaults are those of \cref{sec:expdetails}.
 
 \subsection{Shared pipeline components: head budgets and final refit}
 \label{sec:app-abl-shared}
 
 Both components are inherited from attention matching \citep{zweiger2026fast} and applied to every calibrated method, so their effect is measured per method. \Cref{tab:abl-refit-budget} switches them independently for our methods and the attention-matching baselines; the ported eviction methods are walked through the same switches in \cref{tab:abl-regime}.
 
 \textbf{Head budgets.} The ablation replaces the optimized per-head budgets of \cref{sec:expdetails} by uniform per-layer budgets at the same global ratio. It is the largest single lever, and it is not specific to \textsc{KV-Surgeon}: with the refit in place, \textsc{KV-Surgeon} loses $6.6$ points, AM-HAK $6.9$, and LOO $7.8$. Under uniform budgets \textsc{KV-Surgeon} remains $2.4$ points ahead of LOO and $3.0$ ahead of AM-HAK, so the selection-rule margins of \cref{sec:exp-obs} do not depend on the optimized budgets. KVzip, the one baseline that sets its own head allocation from a global attention threshold, leads to the same conclusion: ported with the shared budgets and refit, it collapses under strong compaction, a median of $15.3$ points below \textsc{KV-Surgeon}-fast at $50\times$ and $100\times$, yet sits a median of $1.6$ points below the ceiling at $5\times$ (\cref{sec:app-grids}); the deficit tracks how hard the compaction is, not how good the score is.
 
 \textbf{Final refit.} The refit is ablated on the LOO support, where it can be switched off without touching selection (\cref{tab:ablation-main}). Serving the selected keys with their original multipliers and values, which is plain eviction on the score of \labelcref{eq:loo}, costs $5.4$ points against LOO with the refit. Across methods the two switches compound: the budgets are the larger lever, worth $+4$ to $+8$ points on top of the refit, and the refit adds $+1$ to $+7$ on top of raw compaction (\cref{tab:abl-refit-budget,tab:abl-regime}). The improvements are not the same for all methods, but \textsc{KV-Surgeon} is ahead in all three settings. Panel (c) of \cref{tab:ablation-pool} decomposes the refit into its two solves on the \textsc{KV-Surgeon} support; it is kept as supporting evidence for \cref{sec:mass} and is not built on elsewhere.
 
 \subsection{KV-Surgeon-specific component: compensation}
 \label{sec:app-abl-comp}
 
 Panel (b) of \cref{tab:ablation-pool} isolates compensation. The surrogate, the $4t$ candidate pool, and the final refit are held fixed and only the compensating update of \labelcref{eq:obs} is disabled, so that elimination proceeds greedily on the same surrogate saliencies. This costs $2.9\pm0.8$ points (seed-paired; $3.0$ between the seed means of \cref{tab:comp-ablation}), and leaves greedy elimination no better than LOO selection on the same articles ($64.7\%$). Two controls show that the gain is not a better standalone objective. Greedy elimination on the exact set loss of \labelcref{eq:family} instead of the surrogate reaches $65.8\%$, $1.1\pm1.3$ below \textsc{KV-Surgeon} and too noisy on $25$ articles to separate from either side. Shrinking the candidate pool to the support itself (pool factor $1$ in panel (a)), which leaves the elimination nothing to eliminate or compensate, costs the same $2.9$ points. The loss-level counterpart, that the compensated support moves toward an exact-refit oracle and that the compensation generalizes to held-out queries, is in \cref{sec:app-surrogate}.
 
 \subsection{Efficiency and design choices}
 \label{sec:app-abl-design}
 
 The remaining choices change accuracy by at most one standard error at their defaults and mainly control cost.
 
 \textbf{Mass term.} Setting $\gamma_0=0$ removes the mass term from the surrogate (\cref{tab:ablation-main}). Accuracy moves by $+0.2$ points, within one standard error, at $2$ s more pipeline time.
 
 \textbf{Candidate pool.} The elimination runs on the $\min\{T,\lceil 4t\rceil\}$ best entries per head by the score of \labelcref{eq:loo} (panel (a) of \cref{tab:ablation-pool}). Widening the pool to the full cache changes accuracy by $0.2$ points ($66.9$ against $66.7$) and multiplies compaction time by $3.8$ ($10$ against $38$ s); halving it to $2t$ costs $1.0$ point.
 
 \textbf{Mild-compaction fallback.} Heads whose budget $t$ exceeds $30\%$ of their cache are served by plain eviction on the score of \labelcref{eq:loo}, without pool, elimination, or final refit: eviction is near-lossless there, the $O(t^3)$ value solve is the dominant cost and is largest exactly for these heads, and for $t>N$ the value solve of \labelcref{eq:finalvalues} is underdetermined. Raising the threshold to $1.0$ disables the fallback (\cref{tab:ablation-main}). The accuracy of \textsc{KV-Surgeon}-fast is unchanged to the reported precision, and its pipeline time rises from $5.4$ to $7.4$ s, so the fallback saves $27\%$ of the pipeline. The fallback was not ablated on the self-study tier.
 
 \textbf{Surrogate refresh.} The frozen surrogate can be re-solved at the current normalization ratios $\mu$ every $256$ or $64$ eliminations, either reweighting the residuals of \labelcref{eq:split} (panel (a) of \cref{tab:ablation-pool}) or consistently, holding that mass fixed so that the refreshed objective agrees with the served output (panel (b)). At the default pool the reweighted refresh costs $1.6$ and $1.9$ points at $1.3\times$ and $2.2\times$ the compaction time, and the consistent refresh costs $1.3\pm0.7$ and $0.7\pm0.2$ points. Across the pool sweep the refreshed arm is never better than the frozen one beyond one standard error of the difference (the largest apparent gain is $+1.0$ at pool factor $1$, with standard errors of $1.0$ and $0.7$), while its cost grows to $11\times$ at the full cache. At the loss level, measured as in \cref{sec:app-surrogate}, the refresh changes which positions survive, Jaccard $0.55$ to $0.85$ against the frozen support, without changing the exact loss after the final refit on calibration or held-out queries (within $\pm2\%$ on every head where a refresh occurs). Freezing is therefore kept for cost, not because it regularizes.
 
 \begin{table}[p]
   \centering
   \footnotesize
   \setlength{\tabcolsep}{4pt}
   \begin{tabular}{@{}llcccc@{}}
     \toprule
     & & \multicolumn{2}{c}{Default} & \multicolumn{2}{c}{Ablated} \\
     \cmidrule(lr){3-4} \cmidrule(l){5-6}
     Ablation & Method & Acc.\ (\%) & Time (s) & Acc.\ (\%) & Time (s) \\
     \midrule
     \multicolumn{6}{@{}l}{\textit{Final refit off (plain eviction on the selected support)}} \\
     & LOO & 65.6\se{0.2} & 38 & 60.2\se{0.4} & 35 \\
     \addlinespace[2pt]
     \multicolumn{6}{@{}l}{\textit{Mass term off ($\gamma_0=0$)}} \\
     & \textsc{KV-Surgeon} & 66.8\se{0.3} & 41 & 67.0\se{0.4} & 43 \\
     \addlinespace[2pt]
     \multicolumn{6}{@{}l}{\textit{Mild-compaction fallback off (threshold $0.3\to1.0$)}} \\
     & \textsc{KV-Surgeon}-fast & 65.5\se{0.2} & 5.4 & 65.5\se{0.2} & 7.4 \\
     \bottomrule
   \end{tabular}
   \caption{One-component ablations on Qwen3-4B with QuALITY at $50\times$, all $50$ articles, five seeds, one standard error over seeds. Time is pipeline time per context (calibration-query generation plus compaction). The self-study controls are the runs of \cref{tab:main-acc}; the fast-tier control is the fast configuration in use at the time of the ablation and is read only against its own row. The head-budget ablation is in \cref{tab:abl-refit-budget}.}
   \label{tab:ablation-main}
 \end{table}
 
 \begin{table}[p]
   \centering
   \footnotesize
   \setlength{\tabcolsep}{4pt}
   \textit{(a) Candidate-pool width and reweighted surrogate refresh (\textsc{KV-Surgeon})}\\[3pt]
   \begin{tabular}{@{}lcccccc@{}}
     \toprule
     & \multicolumn{2}{c}{Refresh off} & \multicolumn{2}{c}{Refresh every 256} & \multicolumn{2}{c}{Refresh every 64} \\
     \cmidrule(lr){2-3} \cmidrule(lr){4-5} \cmidrule(l){6-7}
     Pool factor & Acc.\ (\%) & s & Acc.\ (\%) & s & Acc.\ (\%) & s \\
     \midrule
     $1$ (LOO selection) & 64.0\se{1.0} & 7.9 & 65.0\se{0.7} & 7.8 & 64.1\se{0.6} & 8.0 \\
     $2$ & 65.9\se{1.0} & 8.7 & 66.0\se{0.5} & 9.2 & 65.8\se{0.7} & 10 \\
     $4$ (default) & 66.9\se{0.3} & 10 & 65.3\se{0.3} & 13 & 65.0\se{0.5} & 22 \\
     $8$ & 66.5\se{0.7} & 15 & 65.5\se{0.4} & 28 & 66.0\se{0.3} & 74 \\
     $16$ & 66.8\se{0.4} & 22 & 66.0\se{0.7} & 59 & 64.5\se{0.5} & 181 \\
     $50$ (full cache) & 66.7\se{0.5} & 38 & 65.9\se{0.4} & 126 & 66.7\se{0.5} & 416 \\
     \bottomrule
   \end{tabular}\\[8pt]
   \textit{(b) Selection objective at pool factor $4$, refresh off, same final refit}\\[3pt]
   \begin{tabular}{@{}lcc@{}}
     \toprule
     Variant & Acc.\ (\%) & $\Delta$ vs.\ \textsc{KV-Surgeon} \\
     \midrule
     surrogate with compensation (\textsc{KV-Surgeon}) & 66.9\se{0.6} & --- \\
     surrogate, greedy, no compensation & 63.9\se{1.6} & $-2.9$\se{0.8} \\
     exact loss \labelcref{eq:family}, greedy, no compensation & 65.8\se{2.8} & $-1.1$\se{1.3} \\
     LOO selection (same articles) & 64.7\se{0.9} & $-2.1$\se{0.4} \\
     consistent refresh every 256 & 65.6\se{1.5} & $-1.3$\se{0.7} \\
     consistent refresh every 64 & 66.1\se{0.4} & $-0.7$\se{0.2} \\
     \bottomrule
   \end{tabular}\\[8pt]
   \textit{(c) What the final refit re-solves on the \textsc{KV-Surgeon} support, pool factor $4$}\\[3pt]
   \begin{tabular}{@{}lcccc@{}}
     \toprule
     & \multicolumn{2}{c}{Refresh off} & \multicolumn{2}{c}{Refresh every 256} \\
     \cmidrule(lr){2-3} \cmidrule(l){4-5}
     Final refit re-solves & Acc.\ (\%) & s & Acc.\ (\%) & s \\
     \midrule
     values and multipliers (default) & 66.9\se{0.3} & 10 & 65.3\se{0.3} & 13 \\
     values only, multipliers from the elimination & 65.5\se{0.5} & 9.7 & 64.2\se{0.5} & 12 \\
     multipliers only, values from the elimination & 52.7\se{0.8} & 12 & 42.7\se{1.6} & 12 \\
     none, both from the elimination & 64.7\se{1.1} & 8.5 & 60.9\se{0.5} & 11 \\
     \bottomrule
   \end{tabular}
   \caption{Ablations under the $25$-article protocol on Qwen3-4B with QuALITY at $50\times$, five seeds, one standard error over seeds; accuracies are compared only within this table. The three panels are separate runs, so the same configuration can appear in more than one panel with slightly different numbers. Time is compaction only. (a) The pool holds the $\lceil\text{pool factor}\times t\rceil$ best entries per head by the score of \labelcref{eq:loo}; pool factor $1$ is LOO selection with the shared final refit and $50$ saturates at the full cache. The reweighted refresh re-solves the surrogate at the current normalization ratios $\mu$ every $256$ or $64$ eliminations. (b) Greedy rows eliminate on the stated objective without the compensating update of \labelcref{eq:obs}; the exact-loss row evaluates \labelcref{eq:family} on the current support at every step; the consistent refresh re-solves the surrogate at the current normalization ratios $\mu$ while holding that mass fixed, so that the refreshed objective agrees with the served output. Differences are seed-paired; run times were not recorded for this panel. (c) Whatever the final refit does not re-solve is kept from the coefficients $B$ that the elimination itself leaves on the final support.}
   \label{tab:ablation-pool}
 \end{table}
 
 \begin{table}[p]
 \centering
 \caption{The two shared pipeline components switched independently, at $50\times$ on Qwen3-4B / QuALITY, $50$ articles, five seeds: the support-fixed refit (values and multipliers) and the optimized per-head budgets. All methods use our calibration queries; the AM-OMP value reconstruction cannot be disabled. LOO with optimized budgets and no refit is the plain-eviction row of \cref{tab:ablation-main}; the ported eviction methods are in \cref{tab:abl-regime}.}
 \label{tab:abl-refit-budget}
 \small
 \begin{tabular}{@{}llccc@{}}
 \toprule
  & & \multicolumn{2}{c}{uniform budgets} & optimized budgets \\
 \cmidrule(lr){3-4}
 Method & Queries & no refit & refit & + refit \\
 \midrule
 AM-HAK     & self-study & $54.1\pm0.2$ & $57.2\pm0.5$ & $64.1\pm0.9$ \\
 LOO        & self-study & $53.2\pm0.5$ & $57.8\pm0.6$ & $65.6\pm0.2$ \\
 \textsc{KV-Surgeon} & self-study & $58.2\pm0.4$ & $60.2\pm0.3$ & $66.8\pm0.3$ \\
 AM-OMP     & self-study & -- & $59.0\pm0.5$ & $66.4\pm0.2$ \\
 \midrule
 AM-HAK-fast     & repeat & $48.1\pm0.5$ & $53.2\pm0.3$ & $58.1\pm0.5$ \\
 \textsc{KV-Surgeon}-fast & repeat & $56.3\pm0.5$ & $58.6\pm0.5$ & $65.9\pm0.4$ \\
 AM-OMP-CS     & self-study & -- & $58.1\pm0.4$ & $65.7\pm0.4$ \\
 \bottomrule
 \end{tabular}
 \end{table}
 \clearpage
 
 \section{Compensation at the loss level and calibration robustness}
 \label{sec:app-surrogate}
 
 This section measures compensation directly on the attention-output loss rather than through accuracy, and then asks why repeat-pass calibration costs \textsc{KV-Surgeon} less than it costs the exact single-deletion score; \cref{sec:exp-obs,sec:exp-fast} cite the results. All measurements use Qwen3-4B and Llama-3.1-8B, five QuALITY articles, and every head, that is, $1440$ head--article pairs on Qwen3-4B and $1280$ on Llama-3.1-8B. Scores are computed from the self-study calibration queries, and the exact loss of \labelcref{eq:family} is evaluated on held-out benchmark questions the scores never saw; where the same measurement on the calibration queries differs, it is given in parentheses.
 
 \paragraph{Compensation moves the support toward the oracle and generalizes.} Against an oracle that runs the same greedy backward elimination but scores every candidate deletion by the exact loss \labelcref{eq:family} after refitting the remaining entries, on the $546$ heads where this is affordable ($t\leq57$, Qwen3-4B), LOO selection sits $46\%$ above the oracle's loss and \textsc{KV-Surgeon} $25\%$ (median relative loss $0.213$, $0.183$, and $0.146$; random selection $0.243$), closing $45\%$ of the gap, and on the typical head the closed fraction is $39\%$; the oracle is itself greedy, so these gaps are to exact-loss greedy elimination rather than to the optimal support. \textsc{KV-Surgeon} beats LOO on $84\%$ of heads and still shares only $44\%$ of its keys with the oracle set, because many near-equivalent supports exist. On the supports \textsc{KV-Surgeon} selects at $50\times$, the final refit leaves $25\%$ of the raw-eviction loss on Qwen and $7.5\%$ on Llama on held-out questions, against $25\%$ and $7.7\%$ on the calibration queries: no head overfits, and on no head is the stored cache worse than eviction. On Llama raw eviction at $50\times$ leaves a loss larger than the output itself, so the final refit, which every calibrated method shares, removes most of the loss there, and selection decides the remainder. The coefficients the elimination itself produces already achieve most of this ($31\%$ and $8\%$), but they cannot be served raw: on $1\%$ of Qwen heads and $5\%$ of Llama heads they exceed the eviction loss, and the final refit never does, which is the loss-level counterpart of the final-refit ablation of \cref{sec:app-abl-shared}.
 
 \paragraph{Why repeat-pass calibration costs \textsc{KV-Surgeon} less than the exact score.} Averaged over the eight combinations at $50\times$, replacing self-study queries with a single repeat pass costs \textsc{KV-Surgeon} $1.2$ points (\cref{tab:main-acc}); the drop is $4.3$ points on LongHealth with Gemma-3-12B and at most $1.3$ elsewhere. At the level of rankings, recomputing the scores from a single repeat pass drops the Spearman correlation between the surrogate and the exact loss over random deletion sets, computed within one head and one set size, from $0.905$ to $0.786$ on Qwen and from $0.929$ to $0.833$ on Llama, and that of the score of \labelcref{eq:loo} from $0.881$ to $0.690$ on both, so under cheap queries the surrogate loses about half the ranking fidelity the exact score loses. Cross-head comparability, which is what the per-head budget allocation consumes, is unaffected ($0.81$ and $0.91$ at $98\%$ deletion), so budget allocation does not need self-study queries. The surrogate is therefore less sensitive to the calibration queries than the exact score, not insensitive.
 
 \emph{Hypothesis, not tested as a mechanism.} The factor $(1-P_{ij})^{-2}$ in the exact deletion score amplifies exactly the queries a repeat pass gets wrong: a query for which one token receives unusually large attention contributes disproportionately to that token's score, so an atypical repeat-pass attention pattern moves the exact ranking more than it moves the frozen surrogate, which contains no such amplification. This is consistent with the measurements above and is a candidate mechanism for the fast-tier robustness of \cref{sec:exp-fast}, but it does not establish why AM-HAK degrades more strongly under the same change: AM-HAK ranks tokens by attention mass rather than by either of these objectives, so its larger calibration drop may have a different cause.

 \clearpage
 
 \section{Ported baselines: from the published regime to ours}
 \label{app:abl-regime}
 
 H2O, OBCache, KVMerger, and KVzip are serving-time or self-allocating methods; this section describes how each is ported into the offline pipeline. Each port replaces the method's own queries by the shared repeat-pass calibration queries, keeps its selection score, and serves the selected support with the shared per-head budgets and final refit of \cref{sec:refit}. For OBCache, the published observation window of the last $64$ prompt tokens is replaced by the calibration queries, and its second-order score is aggregated over them. \Cref{tab:abl-regime} validates the port by walking each method from its published configuration to our pipeline one component at a time, at $50\times$ on Qwen3-4B with QuALITY under the $50$-article protocol of \cref{sec:expdetails}.
 
 \begin{table}[h]
 \centering
 \caption{Each external baseline from its published configuration to our pipeline, one component at a time, at $50\times$ on Qwen3-4B / QuALITY. As published: own queries (H2O and KVMerger use context prefill, OBCache its last-64 observation window, KVzip its repeat prompt with a global cross-head threshold), uniform or own head allocation, no refit. The remaining columns use repeat-pass calibration queries.}
 \label{tab:abl-regime}
 \small
 \begin{tabular}{@{}lcccc@{}}
 \toprule
 Method & as published & calib. queries & + refit & + budgets \\
 \midrule
 H2O       & $47.4\pm0.5$ & $48.7\pm0.4$ & $56.1\pm0.4$ & $62.3\pm0.4$ \\
 OBCache   & $43.2\pm0.2$ & $53.0\pm0.4$ & $58.1\pm0.3$ & $62.4\pm0.7$ \\
 KVMerger  & $47.7\pm0.5$ & $47.6\pm0.2$ & $51.5\pm0.5$ & $57.2\pm0.5$ \\
 KVzip     & $36.4\pm0.3$ & $42.6\pm0.3$ & $43.5\pm0.6$ & $49.8\pm0.4$ \\
 \bottomrule
 \end{tabular}
 \end{table}
 
 The pipeline we use, inherited from \citet{zweiger2026fast}, is worth $9$ to $19$ points to each of the eviction methods we ported to the offline compaction regime; the budgets are the largest single step for KVMerger and KVzip, the refit for H2O, and the calibration queries for OBCache. Even with every component, the best ported rule, OBCache at $62.4$, stays $3.5$ points below \textsc{KV-Surgeon}-fast ($65.9$) at the same cost, and H2O at $62.3$ is within noise of OBCache. The two hybrid configurations with their own queries but our budgets (H2O $58.7\pm0.2$, KVMerger $57.5\pm0.5$) confirm that budgets alone do not close the gap.

 \clearpage
 
 \section{Cache construction and the role of mass matching}
 \label{sec:construction}
 
 \textbf{Chunked contexts.} Long structured contexts, for example records of several patients, are compacted one chunk at a time. At inference the compacted blocks sit side by side in one cache, and the softmax normalizes across all of them. Block $b$ therefore competes for the attention of query $i$ through its total unnormalized mass $m_i^b = \sum_{j\in S_b} w_j e^{Z_{ij}}$. A compaction that shrinks a block's mass re-routes attention away from that block and toward its neighbors, even when the reconstruction within the block is perfect. In this setting the absolute mass of a block is meaningful, and it must be preserved alongside the outputs. \Cref{lem:mixed} makes this precise, and \cref{sec:mass} discusses what mass matching can and cannot tell selection.

\begin{lemma}[Serving next to other tokens]
\label{lem:mixed}
Suppose query $i$ attends to the compacted cache together with further tokens whose unnormalized attention sums to $c_i\sum_{\ell=1}^{T}e^{Z_{i\ell}}$ for some $c_i\geq0$ and whose output on their own is $o_i'$. Then the served output is
\begin{equation}
    \frac{\mu_i\,\widehat o_i+c_i\,o_i'}{\mu_i+c_i},
    \qquad\text{compared with}\qquad
    \frac{o_i+c_i\,o_i'}{1+c_i}
    \label{eq:mixed}
\end{equation}
for the original cache. The two agree for every $c_i\geq0$ and every $o_i'$ if and only if $\widehat o_i=o_i$ and $\mu_i=1$.
\end{lemma}
\noindent\textit{Proof.}
The softmax over all tokens has the denominator $\sum_{j\in S}w_je^{Z_{ij}}+c_i\sum_\ell e^{Z_{i\ell}}$ and the numerator $\sum_{j\in S}w_je^{Z_{ij}}\widehat v_j+c_io_i'\sum_\ell e^{Z_{i\ell}}$. Dividing both by $\sum_\ell e^{Z_{i\ell}}$ and using $\sum_{j\in S}w_jP_{ij}=\mu_i$ and $\sum_{j\in S}w_jP_{ij}\widehat v_j=\mu_i\widehat o_i$ from \labelcref{eq:compactedattention,eq:mu} gives the first expression. The original cache is the case $S=\{1,\ldots,T\}$, $w=\one$, $\widehat V=V$, for which $\mu_i=1$ and $\widehat o_i=o_i$. If $\widehat o_i=o_i$ and $\mu_i=1$, the two expressions coincide. Conversely, $c_i=0$ gives $\widehat o_i=o_i$; then both expressions are convex combinations of $o_i$ and $o_i'$ with the weights $\mu_i/(\mu_i+c_i)$ and $1/(1+c_i)$ on $o_i$, which agree for $c_i>0$ and $o_i'\neq o_i$ only if $\mu_i=1$.
\hfill$\square$

 \subsection{The role and limits of mass matching}
 \label{sec:mass}
 
 Attention matching \citep{zweiger2026fast}, the strongest prior offline method, builds its selection and refit around reproducing each query's attention \emph{mass} on the compacted block. Two consequences of \cref{sec:surrogate} matter here: for how the refit must be run, and for what mass matching can tell selection.
 
 \textbf{Per-position mass fitting is harmful alone and misleading as a selection signal.} Non-uniform multipliers change the retained attention pattern, so the original values no longer reconstruct $o_i$ under the reweighted pattern. Fitting multipliers to mass targets without the subsequent value solve is therefore worse than no refit at all, which is why the refit of \cref{sec:refit} applies the two solves only as an ordered unit. This is measured on the \textsc{KV-Surgeon} support: re-solving the multipliers alone scores $12.0$ points below serving the elimination's own coefficients with no refit at all, and $14.2$ below the full final refit (panel (c) of \cref{tab:ablation-pool}). The mechanism is the reweighting rather than the mass objective in particular, so the ablation bounds what mass fitting costs when it is run alone and does not separate it from other multiplier objectives. Mass matching also misleads selection. Weighting the mass term heavily, formally $\gamma\to\infty$ in \labelcref{eq:decoupled}, reduces the saliency \labelcref{eq:saliency} to its mass component, so that entries are ranked by their contribution to reproducing the attention mass alone, as in attention matching, and their values play no role in selection. This discards both factors that the exact loss \labelcref{eq:loo} says matter, the value novelty $\|v_j - o_i\|^2$ and the heavy-hitter weight $(1-P_{ij})^{-2}$. At mild compaction the difference is small, because retained masses stay near one and all criteria agree. At aggressive compaction the two criteria part ways, and the exact loss sides with the output: replacing mass ranking by \textsc{KV-Surgeon} selection inside an otherwise identical pipeline is worth $1.3$ points at $20\times$, $3.0$ at $50\times$, and $4.3$ at $100\times$ (\cref{fig:mean-ratio}).
 
 None of this contradicts the published attention-matching results, because all of those results combine mass matching with a value refit.

\textbf{Served multipliers in chunked caches.} 
\Cref{lem:mixed} says that a compacted block must preserve its attention mass when it is served next to other tokens. We test this on LongHealth with \textsc{KV-Surgeon}-fast on Qwen3-4B-Instruct-2507, where every context holds five patient records compacted as separate chunks. We vary only what multipliers the compacted entries are served with: the box-constrained refit of \cref{sec:refit}, the multipliers that the elimination itself leaves on the support, clamped to the same box, or unit multipliers. Values are refit in all three cases. Each setting is run in two regimes. In the first, all five records are compacted. In the second, two or three records are kept uncompacted next to the compacted ones; the two complementary masks $\{1,3\}$ and $\{0,2,4\}$ are both run, so every record is served once compacted and once uncompacted, and accuracy is split by whether a question asks about a compacted or an uncompacted record.

\begin{table}[h]
  \centering
  \caption{Accuracy (\%) by served multipliers, \textsc{KV-Surgeon}-fast on LongHealth with Qwen3-4B-Instruct-2507, five seeds, standard error over seeds in \textcolor{black!55}{grey}. \emph{All compacted}: all five records compacted. \emph{Mixed}: two or three records served uncompacted; accuracy on questions about the compacted and the uncompacted records is reported separately. Each column covers all $400$ questions.}
  \vspace{9pt}
  \small
  \setlength{\tabcolsep}{5pt}
  \begin{tabular}{@{}lcccccc@{}}
    \toprule
    & \multicolumn{3}{c}{$100\times$} & \multicolumn{3}{c}{$50\times$} \\
    \cmidrule(lr){2-4} \cmidrule(lr){5-7}
    & All & Mixed & Mixed & All & Mixed & Mixed \\
    Multipliers & compacted & compacted & uncompacted & compacted & compacted & uncompacted \\
    \midrule
    Refit (default)            & 57.5\se{0.8} & 58.0\se{0.7} & 78.2\se{0.6} & 66.9\se{0.5} & 67.9\se{0.6} & 79.4\se{0.5} \\
    From the elimination & 56.7\se{0.4} & 56.5\se{0.7} & 79.2\se{0.5} & 68.5\se{0.7} & 67.8\se{0.3} & 78.7\se{0.5} \\
    Unit                       & 55.9\se{0.6} & 45.8\se{0.8} & 81.3\se{0.2} & 65.2\se{0.6} & 59.0\se{0.9} & 80.6\se{0.5} \\
    \bottomrule
  \end{tabular}
  \label{tab:mass-served}
\end{table}

Three things follow from \cref{tab:mass-served}. First, unit multipliers are harmless when everything is compacted the same but harmful when compacted cache chunks are served next to uncompacted ones. With all records compacted they cost $1.6$ and $1.7$ points. With uncompacted neighbors the same compacted records lose $12.2$ points at $100\times$ and $8.9$ at $50\times$, while the uncompacted neighbors gain $1$ to $3$ points. This is the re-routing of \cref{lem:mixed}: a block whose mass falls below the original hands attention to its neighbors. When every block is compacted, the deficits are similar and largely cancel. Second, mass-preserving multipliers make a block's accuracy independent of its neighbors: for both fitted settings the compacted records score the same whether their neighbors are compacted or not. Third, the multipliers that the elimination leaves on the support, clamped to the box of \labelcref{eq:finalweights}, match the refit within seed noise, so the final multiplier solve can be skipped at no cost. Removing the mass term from the surrogate ($\gamma_0=0$) changes every cell within two standard errors, consistent with \cref{sec:app-abl-design}: mass preservation is carried by the served multipliers, not by the selection.
 
 \section{Multiplier diagnostics for the unconstrained elimination}
 \label{app:abl-gates}
 
 \begin{table}[h]
 \centering
 \caption{Diagnostics of the unconstrained multiplier coefficients $\tilde w\triangleq B_{:,d_v+1}/\gamma$ that the elimination leaves on the final support and of the served multipliers $w$ after the box-constrained refit of \labelcref{eq:finalweights}, per ratio, on Qwen3-4B / QuALITY (seed 0, all 14{,}400 head instances; heads served by fallback or zero budget excluded). Columns: fraction of kept entries with $\tilde w_j<0$ (pooled over heads, and the median head); fraction of heads with at least one negative coefficient; median per-head relative movement $\lVert w-\tilde w\rVert_2/\lVert\tilde w\rVert_2$ and correlation of $w$ against $\tilde w$; fraction of served multipliers pinned at the bounds $e^{\pm\beta}$, $\beta=3$.}
 \label{tab:abl-gates}
 \small
 \begin{tabular}{@{}lcccccc@{}}
 \toprule
 Ratio & $\tilde w<0$ & median head & heads w/ any & rel. move & corr. & at bounds \\
 \midrule
 $100\times$ & $0.42\%$ & $0.00\%$ & $13.6\%$ & $0.81$ & $0.39$ & $1.3\%$ / $14.7\%$ \\
 $50\times$  & $0.33\%$ & $0.00\%$ & $19.4\%$ & $0.70$ & $0.51$ & $1.3\%$ / $9.4\%$ \\
 $20\times$  & $0.24\%$ & $0.00\%$ & $24.6\%$ & $0.57$ & $0.62$ & $1.8\%$ / $5.9\%$ \\
 $10\times$  & $0.24\%$ & $0.00\%$ & $28.0\%$ & $0.54$ & $0.68$ & $2.9\%$ / $5.5\%$ \\
 $5\times$   & $0.28\%$ & $0.00\%$ & $33.0\%$ & $0.58$ & $0.70$ & $4.7\%$ / $4.1\%$ \\
 \bottomrule
 \end{tabular}
 \end{table}
 
 \Cref{tab:abl-gates} shows the impact of a relaxation that makes closed-form compensation possible, i.e., relaxing the multiplier coefficients to real numbers instead of positive numbers. Fewer than half a percent of kept coefficients end negative at any ratio, the median head has no negative coefficients, and the numbers are unchanged between the repeat and self-study query tiers. The refit is calibration, not sign repair: it rescales substantially (median relative movement $0.5$ to $0.8$, correlation $0.4$ to $0.7$) and its pinning at the upper bound grows from $4\%$ of multipliers at $5\times$ to $15\%$ at $100\times$, where restoring the attention mass requires the largest multipliers. Negative coefficients concentrate in early layers (1--6), and refit movement is largest in late layers (25--34).

\end{document}